\documentclass[runningheads]{llncs}

\usepackage{eccv}

\usepackage{eccvabbrv}

\usepackage{graphicx}
\usepackage{booktabs}

\usepackage[accsupp]{axessibility}  % Improves PDF readability for those with disabilities.

\usepackage{hyperref}

\usepackage{orcidlink}

\usepackage{multirow}
\usepackage{pifont}
\usepackage{colortbl}
\usepackage{xcolor}
\usepackage[linesnumbered,ruled,vlined]{algorithm2e}

\SetKwInput{KwInput}{Input}
\SetKwInput{KwOutput}{Output}

\newcommand{\aka}{\textsc{a.k.a.}\@\xspace}
\definecolor{tablehighlightgray}{RGB}{231, 231, 231}

\begin{document}

\title{PanoSAM2: Lightweight Distortion- and Memory-aware Adaptions of SAM2 for 360 Video Object Segmentation} 

\titlerunning{PanoSAM2}

\author{Dingwen Xiao\inst{1}\thanks{Equal contribution. $\dag$Corresponding author.} \and
Weiming Zhang\inst{1}$^{*}$ \and
Shiqi Wen\inst{1} \and
Lin Wang\inst{2}$^{\dag}$}

\authorrunning{Xiao.~Author et al.}

\institute{The Hong Kong University of Science and Technology (Guangzhou)\\
\and
Nanyang Technological University\\
\email{\{dxiaoaf, wzhang915, swen750\}@connect.hkust-gz.edu.cn}\\
\email{linwang@ntu.edu.sg}}

\maketitle

\begin{center}
  \includegraphics[width=\linewidth]{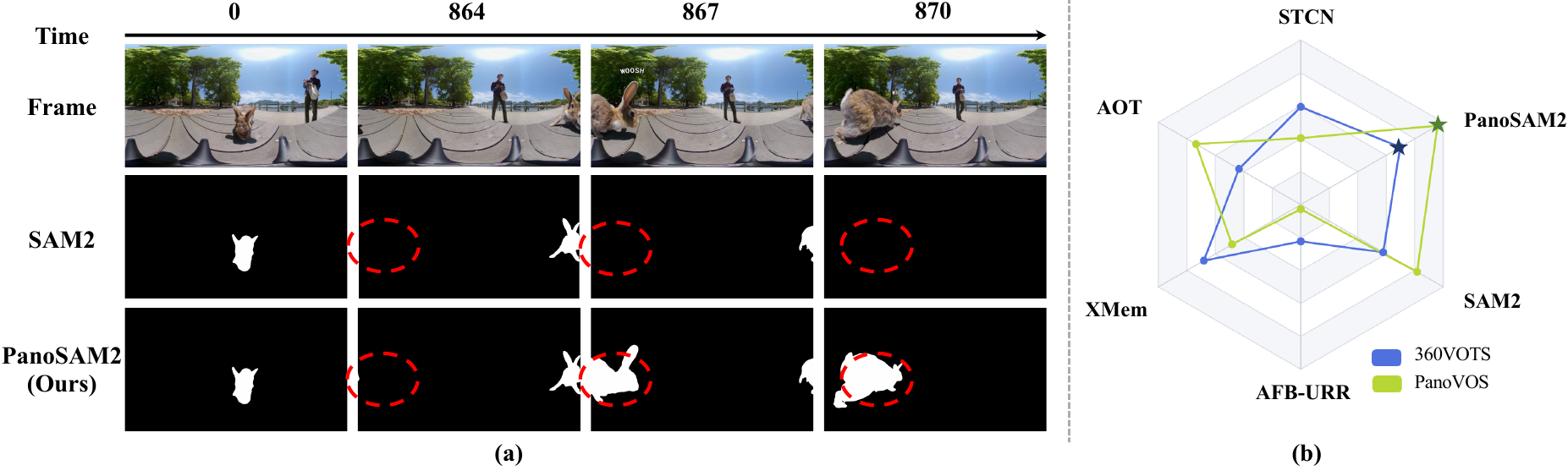}
\end{center}
{\captionof{figure}{Our PanoSAM2 achieves superior results on 360 Video Object Segmentation via spherical distortion and geometry adaptations of SAM2. (a) Sample results from SAM2 and PanoSAM2 on the 360 panoramic video frames, showing how PanoSAM2 better segments the target across different time frames (\textcolor{red}{red dash circle}). (b) Plot comparing the performance of PanoSAM2 against existing methods on 360VOTS \cite{xu2025360vots} and PanoVOS \cite{yan2024panovos} (star icon for the best model), demonstrating its superior video segmentation capabilities.}
\label{fig:teaser_fig}}

\begin{abstract}
  360 video object segmentation (360VOS) aims to predict temporally-consistent masks in 360 videos, offering full-scene coverage, benefiting applications, such as VR/AR and embodied AI. Learning 360VOS model is nontrivial due to the lack of high-quality labeled dataset. Recently, Segment Anything Models (SAMs), especially SAM2 -- with its design of memory module -- shows strong, promptable VOS capability. However, directly using SAM2 for 360VOS yields implausible results as 360 videos suffer from the projection distortion, semantic inconsistency of left-right sides, and sparse object mask information in SAM2's memory. To this end, we propose \textbf{PanoSAM2}, a novel 360VOS framework based on our lightweight distortion- and memory-aware adaptation strategies of SAM2 to achieve reliable 360VOS while retaining SAM2’s user-friendly prompting design. Concretely, to tackle the projection distortion and semantic inconsistency issues, we propose a \textbf{Pano-Aware Decoder} with seam-consistent receptive fields and iterative distortion refinement to maintain continuity across the 0°/360° boundary. Meanwhile, a \textbf{Distortion-Guided Mask Loss} is introduced to weight pixels by distortion magnitude, stressing stretched regions and boundaries. To address the object sparsity issue, we propose a \textbf{Long–Short Memory Module} to maintain a compact long-term object pointer to re-instantiate and align short-term memories, thereby enhancing temporal coherence. Extensive experiments show that PanoSAM2 yields substantial gains over SAM2: +\textbf{5.6} on 360VOTS and +\textbf{6.7} on PanoVOS, showing the effectiveness of our method. % Code will be publicly available.
\end{abstract}

\section{Introduction}
\label{sec:intro}

360 video object segmentation (360VOS) aims to track and segment target objects across 360 videos, given their mask in the first frame. Unlike conventional perspective VOS with a limited field-of-view (FoV), 
360 videos -- with the commonly used projection type of Equirectangular projection (ERP) -- can observe all entire spherical scene, maintaining awareness of objects from all directions. This property makes it particularly valuable for immersive applications such as VR/AR \cite{10.1007/978-3-030-50726-8_3, Jost18082021}, autonomous driving \cite{wen2024panacea, petrovai2022semantic, zhang2024goodsam, yan2024panovos}, and embodied robotics \cite{zhang2018detection, huang2022360vo}, which demand temporally consistent tracking with stable object identity.
% and low latency. 
However, an ERP image, \aka, panorama\footnote{In this paper, ERP and panoramic are interchangeably used.} samples pixels with a higher density at the poles compared to the equator, resulting in spherical distortions. Moreover, the left–right ERP borders correspond to adjacent longitudes, causing the semantic inconsistency at the 0°/360° seam.  
These render the perspective 2D imagery-based VOS research \cite{xu2025segment, yang2020collaborative, liang2020video, yang2021associating, oh2019video, cheng2021rethinking, cheng2022xmem, cheng2024putting} less applicable or effective.
Only recently, a few studies \cite{yan2024panovos, xu2025360vots} consider the geometric and photometric characteristics unique to omnidirectional capture. However, the benchmarks \cite{xu2025360vots, yan2024panovos} for 360VOS remain much under-explored compared to their 2D counterparts \cite{xu2018youtube, pont20172017, 7780454, ravi2024sam}, and their generalization capacity is limited. Bridging this gap requires rethinking both model architectures and optimization objectives to account for the distortion and spherical continuity in 360 imagery.

Recently, Segment Anything Model (SAM), especially SAM2 \cite{ravi2024sam} is a prompt-based VOS foundation model that shows strong zero-shot, promptable capabilities thanks to its user-friendly interface (points, boxes, or masks) and a memory module trained over 50K+ videos. While SAM2 has inspired extensions across domains \cite{zhou2025camsam2, mei2025sam}, directly applying it to 360VOS produces implausible results (see \cref{fig:teaser_fig}) caused by the distortion, left–right semantic inconsistency at the 0°/360° seam, and small masks that often appeared in the target objects, leading to sparse object evidence in SAM2’s memory. Under such sparsity and occlusion, short-term memory can drift or forget, echoing prior observations in memory-based VOS \cite{ding2025sam2long, cheng2022xmem, bekuzarov2023xmem++, khoreva2016learning, yang2021associating, yan2024panovos}.

To address these challenges, we explore a novel idea: lightweight \textbf{distortion- and memory-aware adaptation} that preserves SAM2’s promptable design while making it reliable for panoramic videos. Intuitively, we propose \textbf{PanoSAM2}, a novel 360VOS framework that can achieve robust 360VOS (see \cref{fig:teaser_fig}) with strong generalization via three interconnected technical components.
Firstly, to tackle the spherical continuity and projection distortion, we propose a \textbf{Pano-Aware (PA) Decoder} that reshapes the mask decoding process (see \cref{subsec:pa_decoder}). Concretely, it performs left–right wrap concatenation to build seam-consistent receptive fields, ensuring that features at the right border attend to their true neighbors at the left border (\ie, the 0°/360° seam). Then, the decoder conditions on the previous-frame mask to apply iterative distortion refinement, progressively correcting features near high-distortion latitudes during decoding. This geometry-aware design remarkably \textit{reduces seam breaks and improves boundary fidelity while remaining a lightweight decoder}.

On top, we introduce a \textbf{Distortion-Guided Mask Loss}, a geometry-aware objective that weights pixels by their distortion magnitude under ERP (see \cref{subsec:dgl}). Intuitively, pixels in highly stretched regions (and near boundaries) contribute more to the loss, encouraging projection-robust masks and sharper boundaries. \textit{The loss is simple to compute, architecture-agnostic, and complements the PA Decoder by aligning the optimization target with the ERP}.

To enhance temporal robustness under sparsity and occlusion, we propose a novel  Long–Short Memory Module (LSMM) (see \cref{subsec:lsmm}). It maintains a compact long-term object pointer—an object-level summary distilled from historical observations—that periodically re-instantiates and aligns the short-term memory. By injecting this pointer into memory attention alongside recent features, PanoSAM2 resists drift, rapidly recovers from occlusions, and prevents dominance by frames where the foreground is tiny or absent. \textit{This design stabilizes identity while preserving responsiveness to new inputs, thereby improving temporal coherence in challenging panoramic scenes}.

We evaluate our PanoSAM2 on the 360VOTS \cite{xu2025360vots} and PanoVOS \cite{yan2024panovos}, observing consistent gains over SAM2. In particular, PanoSAM2 improves by +\textbf{5.6} on 360VOTS test set and +\textbf{6.7} on PanoVOS validation set, indicating that panoramic geometry and memory constraints can be effectively addressed without discarding the promptable interface. Ablation studies attribute the improvements to three innovative components. Notably, these benefits come with modest overhead, preserving the efficiency for practical applications.

In summary, our contributions are \textbf{four-fold}: 
\begin{itemize}
    \item We make the \textbf{first} attempt to leverage SAM2’s zero-shot, prompt-based paradigm and propose PanoSAM2, a novel approach for the challenging 360VOS task by introducing a tightly integrated, architecture-specific design.
    \item We propose the Pano-Aware Decoder that enforces left–right semantic continuity and mitigates ERP distortion; a Distortion-Guided Mask Loss that aligns optimization with spherical sampling; 
    \item We propose the Long–Short Memory Module that couples long-term object pointers with short-term memory to prevent drift under sparsity and occlusion. 
    \item We show that PanoSAM2 shows state-of-the-art (SoTA) performance on diverse benchmark 360VOS datasets ($\mathcal{J}\&\mathcal{F}$ score of \textbf{65.8} on 360VOTS test set and \textbf{78.1} on PanoVOS validation set) and strong generalization capabilities. Notably, the proposed architecture exhibits strong cross-dataset generalization, revealing principles for robust 360VOS.
\end{itemize}

\section{Related Works}
\label{sec:related_work}

\subsection{360 Video Object Segmentation.} 
Compared with perspective settings, object tracking and segmentation in panoramic videos remain underexplored. PanoVOS \cite{yan2024panovos} introduced the 360VOS task, releasing a dedicated dataset (PanoVOS) and the PSCFormer baseline. PSCFormer addresses equirectangular projection (ERP) distortion and semantic consistency by applying left–right padding to preserve wrap-around continuity and by restricting pixel-level attention windows, thereby reducing cross-sphere confusion while maintaining efficiency. Besides, 360VOTS \cite{xu2025360vots} proposes a Bounding Field-of-View (BFoV) mechanism that handcrafts the next-frame search region from the previous prediction, effectively “windowing” the panorama so that conventional VOS models can be used without architectural changes. While BFoV enables plug-and-play reuse of mature 2D pipelines, it introduces a runtime bottleneck due to sequential localized searches and is brittle when targets are heavily occluded, leave the window, or re-enter with large appearance or viewpoint shifts. We argue that methods tailored to ERP must simultaneously handle severe projection distortion, seam consistency, and sparsely distributed target pixels, challenges that are only partially addressed by padding and local attention. To this end, \textit{we propose PanoSAM2 that explicitly incorporates geometry-aware decoding and memory while retaining SAM2 capabilities for panoramic streams.}

\subsection{SAM-Based Video Object Segmentation}
The SAM family \cite{kirillov2023segment, ravi2024sam} established segmentation foundation models whose promptable design enables broad transfer. Building on this, several works \cite{rajivc2025segment, mei2025sam, zhou2025camsam2, zhang2025leader360v, liu2024surgical, mendoncca2025seg2track} adapt it to video. SAM-PT \cite{rajivc2025segment} couples a point-tracking model with SAM: tracked points on the target are fed as prompts to segment each frame, requiring no task-specific training. SAM-I2V \cite{mei2025sam} synthesizes prompts from temporal cues by combining past masks and current-frame features, then performs inference via SAM’s prompt encoder and mask decoder. SAM2Long \cite{ding2025sam2long} observes track drift under occlusion, and models mask uncertainty by exploring a tree of hypotheses, selecting favorable branches; the analysis further highlights the value of long-term memory for stable segmentation. CamSAM2 \cite{zhou2025camsam2} extends SAM2 to camouflaged object tracking, injecting camouflage tokens to derive object prototypes that correct current predictions. MAPS \cite{yang2025maps} explores the effect of adding more representative frames to memory. While effective in perspective settings, \textit{these approaches struggle in 360 video: affine priors become unstable across ERP distorted regions, and traditional receptive areas cannot handle semantic consistency for left and right boundaries. Consequently, direct application to panoramic streams yields degraded performance.} In contrast, our designs effectively fit the characteristics of 360 video and are integrated into SAM2’s architecture, going beyond direct adaptation of existing 360 strategies.

\begin{figure}[tb]
    \centering
    \includegraphics[width=0.99\textwidth]{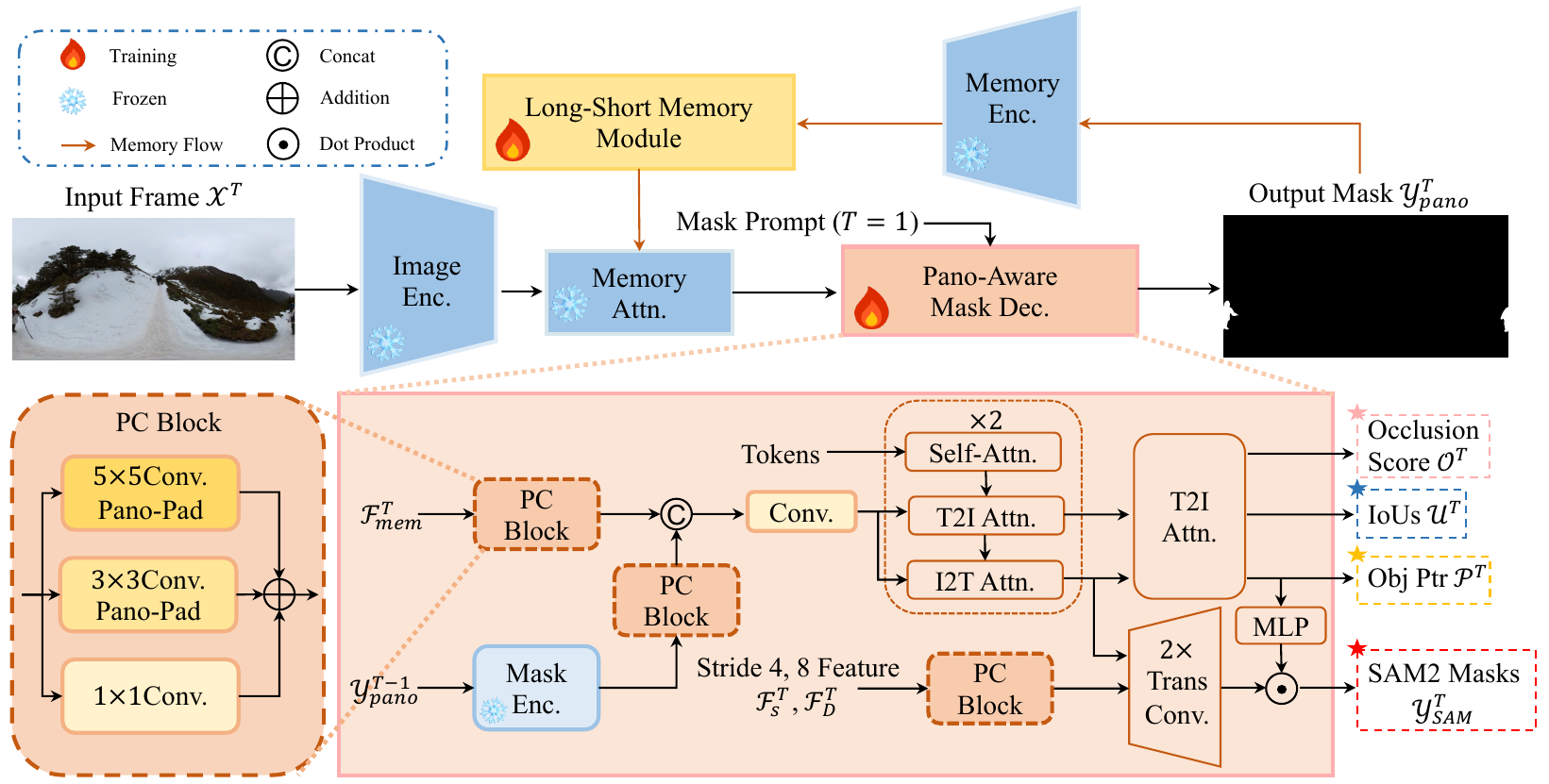}
    \caption{\textbf{Overview of our PanoSAM2 framework}. Compared with SAM2 \cite{ravi2024sam}, it has two architectural contributions: Pano-Aware (PA) Decoder and Long-Short Memory Module (LSMM).
    }
    \label{fig:panosam2}
\end{figure}

\section{Methodology}
\label{sec:methodology}
\noindent \textbf{Preliminaries: SAM2.}
\label{subsec:sam2}
SAM2 \cite{ravi2024sam} is a prompt-based foundation model for image and video object segmentation. It is trained on SA-1B \cite{kirillov2023segment} and SA-V \cite{ravi2024sam} with over 50K+ videos and exhibits strong zero-shot capacity, allowing evaluation on unseen data without task-specific finetuning.

For the frame $\mathcal{X}^T\in\mathbb{R}^{H\times W\times 3}$ at time $T$, SAM2 input it into the Hiera image encoder \cite{ryali2023hiera} to extract feature and then memory-condition it to $\mathcal{F}^{T}_{mem}\in\mathbb{R}^{\frac{H}{16}\times \frac{W}{16}\times 256}$ by operating cross-attention with memory from the memory bank. If it is the first frame, a prompt (point, box, or mask) is provided and encoded. $\mathcal{F}^{T}_{mem}$ is input into the mask decoder together with prompt information (if possible) to decode three mask logits $\mathcal{Y}^T_{SAM}\in\mathbb{R}^{H\times W\times 3}$, an object pointer $\mathcal{P}^T\in\mathbb{R}^{d_p}$ ($d_p$ for pointer dimension) for object-level information, $\mathcal{U}^T\in\mathbb{R}^3$ that predicts the $IoU$ scores of the three predicted masks with ground truth, and an occlusion score $\mathcal{O}^T\in\mathbb{R}$ for the probability of object being visible in this frame. In parallel, the mask with the maximum $IoU$ score and unconditioned frame feature are combined by a memory encoder and, together with the $\mathcal{P}^T$, written into the memory bank. By default, the bank retains only the most recent six frames' memory. 

\noindent \textbf{Our Idea.} However, as SAM2 fails to model equirectangular distortion and left–right seam semantic continuity. Moreover, under heavy distortion or after occlusion, the short memory policy can forget the target or drift to another object, leading to identity breaks. To address these gaps, we propose PanoSAM2 for 360VOS, as depicted in \cref{fig:panosam2}.
We elaborate a \textbf{Pano-Aware (PA) decoder} (see \cref{subsec:pa_decoder}) and \textbf{Distortion-Guided Loss} (see \cref{subsec:dgl}) to tackle the boundary-consistent and distortion-aware prediction.  The loss is simple to compute, architecture-agnostic, and complements the PA Decoder by aligning the optimization target with the ERP.
Then, we articulate the \textbf{Long-Short Memory Module (LSMM)} in \cref{subsec:lsmm} that augments temporal information with additional long-term memory to enhance temporal robustness under sparsity and occlusion.

\subsection{Pano-Aware Decoder}
\label{subsec:pa_decoder}

\noindent \textbf{Insight.}  Our PA decoder adapts SAM2’s mask decoder to 360° FoV by building geometry awareness into the network. Unlike CamSAM2 \cite{zhou2025camsam2} and Seg2Track-SAM2 \cite{mendoncca2025seg2track}, which modify SAM2’s outputs, our design avoids \textbf{splitting seam-spanning objects into two identities by making the decoder itself seam-consistent and distortion-aware}.

In PA Decoder, memory-conditioned features first pass through a Pano-Consistency (PC) block composed of three convolutions with different kernel sizes. For the $3\times3$ and $5\times5$ layers, we apply left–right wrap padding: features at the left border are padded with the right border and vice versa. This PC operation $PC(\cdot)$ is formulated in \cref{eq:pc_block}, where $s$ is for kernel size, $Cat$ is for concatenating at the width dimension, and the padding $p$ is set to 1 and 2 for $s=3$ and $s=5$, respectively. This preserves spatial size while allowing receptive fields to cross the 0°/360° seam, enabling the PA decoder to attend to true spherical neighbors.
\begin{equation}
\small
\label{eq:pc_block}
\begin{split}
  \mathcal{F}^{T}_{mem,l} &= \mathcal{F}^{T}_{mem}[:,:p]
  \\
  \mathcal{F}^{T}_{mem,r} &= \mathcal{F}^{T}_{mem}[:,\frac{W}{16}-p:]
  \\
  PC(\mathcal{F}^{T}_{mem}) &= Conv_s(Cat(\mathcal{F}^{T}_{mem,r}, \mathcal{F}^{T}_{mem}, \mathcal{F}^{T}_{mem,l}))
\end{split}
\end{equation}

For 360 videos, ERP distortion is either present from the first frame or emerges gradually with motion. The first case is guided by the initial prompt mask and needs no extra handling. For gradual changes, we fuse previous-frame mask cues: the last prediction is passed through SAM2’s frozen memory-encoder mask downsampler to obtain mask features, concatenated with the output of the PC block, and fused via a convolution to stabilize features in newly distorted regions. The fused features then undergo multi-round cross-attention with tokens as in SAM2, ensuring consistent refinement across frames. During transpose convolutions, shallow features $\mathcal{F}^T_s$ and $\mathcal{F}^T_d$ from the image encoder are processed by PC Blocks to retain seam consistency and further enhance spatial alignment. Finally, the mask $\mathcal{Y}^T_{pano}$ with the highest $IoU$ score in $\mathcal{Y}^T_{SAM}$ is written to memory via the memory encoder, enabling reliable propagation in subsequent frames.

\begin{figure}[tb]
    \centering
    \includegraphics[width=0.87\textwidth]{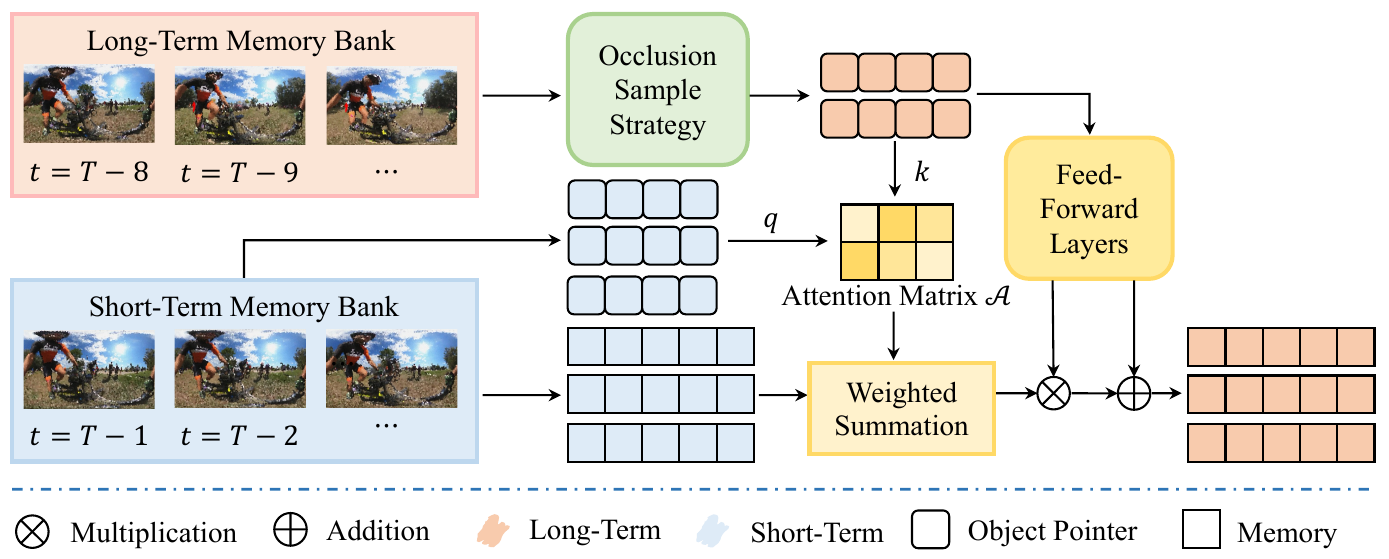}
    \caption{\textbf{Overview of our LSMM framework.}}
    \label{fig:lsm}
    \vspace{-5pt}
\end{figure}

\subsection{Long-Short Memory Module (LSMM)}
\label{subsec:lsmm}
\noindent \textbf{Insight.}  Prior work \cite{ding2025sam2long, cheng2022xmem, bekuzarov2023xmem++, yang2025maps} shows that long-term memory benefits video segmentation, while SAM2 mainly preserves it via iterative updates from the prompted frame. In 360 videos, sparse object visibility further weakens long-range cues. \textbf{We introduce LSMM that fuses long- and short-term information without increasing memory footprint}, as depicted in \cref {fig:lsm}. 

LSMM keeps only object pointers for long-term frames in a bank and drops dense features. An Occlusion Sample Strategy selects key frames by weighted sampling with an occlusion score; pseudocode is provided in the supplementary material. Denote the sampled long-term pointers as \(\mathcal{P}^T_L\in\mathbb{R}^{L\times d_p}\) and short-term pointers as \(\mathcal{P}^T_S\in\mathbb{R}^{6\times d_p}\), We compute an attention matrix \(\mathcal{A}\in\mathbb{R}^{L\times 6}\) that measures long–short similarity and use it to reweight the short-term memory \(\mathcal{M}^T_S\in\mathbb{R}^{6\times \frac{H}{16}\times \frac{W}{16}\times d_m}\), where $d_m$ is the embedding dimension of memory, yielding \(\widetilde{\mathcal{M}}^T_S\).
To inject long-range context, we apply FiLM \cite{perez2018film}: a feed-forward network predicts per-channel scales and biases for the reweighted short-term memory, producing a pseudo long-term memory \(\mathcal{M}^T_L\in\mathbb{R}^{L\times \frac{H}{16}\times \frac{W}{16}\times d_m}\):
\begin{equation}
\label{eq:film}
\begin{split}
\boldsymbol{\gamma},\boldsymbol{\beta} &= \mathrm{FFN}(\mathcal{P}^T_L), \\
\mathcal{M}^T_L &= \widetilde{\mathcal{M}}^T_S \odot \boldsymbol{\gamma} + \boldsymbol{\beta},
\end{split}
\end{equation}
where \(\mathrm{FFN}(\cdot)\) denotes feed-forward network, \(\odot\) is element-wise multiplication, and \(\boldsymbol{\gamma},\boldsymbol{\beta}\in\mathbb{R}^{d_m}\) are broadcast over spatial and short-term dimensions. As in SAM2, we concatenate $\mathcal{M}^T_L$, $\mathcal{M}^T_S$, and pointer sets ($\mathcal{P}^T_L$ and $\mathcal{P}^T_S$), and pass them to the memory attention to condition image features.

\subsection{Distortion-Guided Mask Loss}
\noindent \textbf{Insight.} We propose a distortion-guided loss for 360VOS, while keeping the output head design of SAM2 \cite{ravi2024sam} that outputs three mask logits $\mathcal{Y}_{SAM}^T$, three corresponding $IoU$ scores $\mathcal{U}^T$, and an occlusion score $\mathcal{O}^T$ per frame. The output mask $\mathcal{Y}_{pano}^T$ is the one with the maximum $IoU$ score. \textbf{The key challenge in 360-specific loss design is the combination of projection-induced distortion and severe class imbalance: foreground often occupies tiny regions}.

\begin{figure}[tb]
    \centering
    \includegraphics[width=\textwidth]{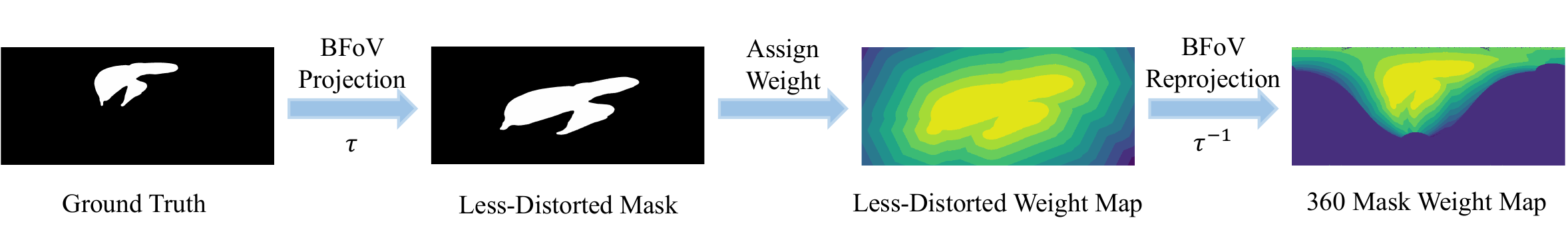}
    \caption{\textbf{Distortion-guided 360 mask weight calculation.}}
    \label{fig:loss}
    \vspace{-5pt}
\end{figure}

To address these, we employ the bounding field-of-view (BFoV) \cite{xu2025360vots} projection $\tau$ for the ground-truth object mask $\mathcal{Y}^{T}_{gt}$ using camera geometry and obtain a less-distorted region inside that BFoV where projection stretching is minimized. As visualized in \cref{fig:loss}, $\tau(\mathcal{Y}^{T}_{gt})$ yields a robust estimate of the foreground proportion, capturing object sparsity for the current frame. Besides, it defines a spatial prior that differentiates reliable evidence from heavily distorted zones. We derive pixel-wise weights $W$ from it. Foreground pixels receive a uniform $w_f$ weight equal to the foreground proportion in $\tau(\mathcal{Y}^{T}_{gt})$, bounded by maximum value $w_{max}$ and minimum value $w_{min}$, encouraging the model to learn from rare positives without overwhelming the objective. Background pixels receive spatially varying weights that are scaled by the complementary proportion and decay with normalized distance to the object boundary, as formulated in \cref{eq:background_weight}, where $D[i,j]$ stands for the distance of pixel $[i,j]$ to the mask boundary and $D_{max}$ is the maximum distance. Thus, hard negatives near the contour are emphasized while far-away background contributes little. At last, we reproject the weight map to the spherical space by $\tau^{-1}$ and fill up the other area with the minimum of $W$. $\tau^{-1}(W)$ is utilized for weight in binary cross-entropy loss  $\mathcal{L}_{BCE}$ for mask optimization.
\begin{equation}
\small
\label{eq:background_weight}
W[i,j]=\frac{1}{w_f}+(w_f-\frac{1}{w_f})\sqrt{1-\frac{D[i,j]}{D_{max}}}
\end{equation}

\label{subsec:dgl}

For the auxiliary heads, we follow SAM2. The $IoU$ score head is trained with mean-squared error against the true $IoU$ value $\mathcal{U}^T_{gt}$ from $\mathcal{Y}^T_{SAM}$ and $\mathcal{Y}^T_{gt}$, and the occlusion score head with binary cross-entropy against the label $\mathcal{O}^T_{gt}$ of whether the object appears in $\mathcal{Y}^T_{gt}$. The overall objective is a weighted summation of the mask, $IoU$, and occlusion terms, as described in \cref{eq:loss}:
\begin{equation}
\label{eq:loss}
\begin{split}
\mathcal{L}_{mask}&=\lambda_{bce}\mathcal{L}_{BCE}(\mathcal{Y}^T_{pano}, \mathcal{Y}^T_{gt}, \tau^{-1}(W)) +\lambda_{dice}\mathcal{L}_{dice}(\mathcal{Y}^T_{pano}, \mathcal{Y}^T_{gt})\\
\mathcal{L}&=\lambda_{IoU}\mathcal{L}_{MSE}(\mathcal{U}^T, \mathcal{U}^T_{gt})+\mathcal{L}_{mask}+\lambda_{occ}\mathcal{L}_{BCE}(\mathcal{O}^T, \mathcal{O}^T_{gt})
\end{split}
\end{equation}
where $\mathcal{L}_{dice}$ refers to the dice loss, and $\mathcal{L}_{MSE}$ represents the mean squared error loss. The parameters $\lambda_{bce}$, $\lambda_{dice}$, $\lambda_{IoU}$, and $\lambda_{occ}$ correspond to the weights applied to the mask BCE loss, mask dice loss, $IoU$ loss, and occlusion score loss, respectively. This design does not alter inference cost, and explicitly aligns supervision with spherical geometry and the statistics of omnidirectional scenes.

\begin{table}
\caption{\textbf{Comparisons between our method and prior arts on the 360VOTS test dataset.} Our PanoSAM2 achieves a new state-of-the-art performance. The best results are shown in \textbf{bold}, and * means the method is re-implemented and reproduced.}
\label{tab:360vots_result}
\centering
\resizebox{0.85\textwidth}{!}{
\begin{tabular}{lcccc|cc|c}
\toprule
\multicolumn{1}{l|}{VOS Tracker}                      & \multicolumn{1}{c|}{Backbone}     & $\mathcal{J}\&\mathcal{F}$          & $\mathcal{J}$             & $\mathcal{F}$ & Memory (GB) & GFLOPs & FPS           \\ \midrule 
\multicolumn{8}{l}{\textit{Perspective-Video-Specified}}\\ \midrule 
\multicolumn{1}{l|}{CFBI \cite{yang2020collaborative}}                             & \multicolumn{1}{c|}{ResNet-101}   & 46.3          & 41.2          & 51.3   & - & - & -       \\
\multicolumn{1}{l|}{CFBI+ \cite{yang2020CFBIP}}                            & \multicolumn{1}{c|}{ResNet-101}   & 48.0          & 42.8          & 53.2 & - & - & -           \\
\multicolumn{1}{l|}{TarVis \cite{athar2023tarvis}}                           & \multicolumn{1}{c|}{Swin-L}       & 36.8          & 32.5          & 41.1 & - & - & -          \\
\multicolumn{1}{l|}{GMVOS \cite{lu2020video}}                           & \multicolumn{1}{c|}{ResNet-50}       & 47.7          & 43.1          & 52.2 & - & - & -          \\
\multicolumn{1}{l|}{UNICORN \cite{yan2022towards}}                           & \multicolumn{1}{c|}{ConvNeXt-L}       & 40.4          & 33.9          & 46.9  & - & - & -         \\
\multicolumn{1}{l|}{AFB-URR \cite{NEURIPS2020_liangVOS}}                          & \multicolumn{1}{c|}{ResNet-50}    & 43.5          & 38.9          & 48.1    & - & - & -       \\
\multicolumn{1}{l|}{STM \cite{oh2019video}}                              & \multicolumn{1}{c|}{ResNet-50}     & 40.1          & 36.4          & 43.8    & - & - & -       \\
\multicolumn{1}{l|}{STCN \cite{cheng2021rethinking}}                             & \multicolumn{1}{c|}{ResNet-50}     & 60.9          & 55.0          & 66.7   & 2.7   & 165.4 & 23.8  \\
\multicolumn{1}{l|}{AOT \cite{yang2021associating}}                              & \multicolumn{1}{c|}{MobileNet-V2} & 53.4          & 47.1          & 59.7 & - & - & -         \\
\multicolumn{1}{l|}{TBD \cite{cho2022tackling}}                              & \multicolumn{1}{c|}{DenseNet-121} & 53.6          & 47.4          & 59.8   & - & - & -       \\
\multicolumn{1}{l|}{RTS \cite{paul2022robust}}                              & \multicolumn{1}{c|}{ResNet-50} & 59.3         & 54.0          &64.5        & - & - & -  \\
\multicolumn{1}{l|}{TBD \cite{mao2021joint}}                              & \multicolumn{1}{c|}{ResNet-50} & 53.7         &48.7          &58.7          & - & - & - \\
\multicolumn{1}{l|}{XMem \cite{cheng2022xmem}}                             & \multicolumn{1}{c|}{ResNet-50}     & 65.0          & 59.6          & 70.3    & 3.5   & 361.2 & 22.5      \\
\multicolumn{1}{l|}{SAM2 \cite{ravi2024sam}}                             & \multicolumn{1}{c|}{Hiera-T}      & 59.4          & 53.9          & 64.9     & 3.6       & 686.7   & 42.6     \\
\multicolumn{1}{l|}{SAM2 \cite{ravi2024sam}}                             & \multicolumn{1}{c|}{Hiera-S}      & 60.2          & 56.8          & 63.6      & 4.1       & 751.4  & 39.3    \\
\multicolumn{1}{l|}{SAM2Long \cite{ding2025sam2long}}                         & \multicolumn{1}{c|}{Hiera-T}      & 59.8          & 54.4          & 65.2  & 4.0 & 735.8 & 29.4        \\
\multicolumn{1}{l|}{SAM2Long \cite{ding2025sam2long}}                         & \multicolumn{1}{c|}{Hiera-S}      & 61.1          & 57.5          & 64.7  & 4.4 & 792.3 & 27.4 \\ \midrule 
\multicolumn{8}{l}{\textit{360-Video-Specified}}\\ \midrule
\multicolumn{1}{l|}{PSCFormer \cite{yan2024panovos}-B*}  & \multicolumn{1}{c|}{ResNet-50}     & 61.0          & 57.7          & 64.3  & 2.8   & 274.7 & 18.5        \\
\multicolumn{1}{l|}{PSCFormer \cite{yan2024panovos}-L*} & \multicolumn{1}{c|}{ResNet-50}     & 62.5          & 58.8          & 66.2  & - & - & -        \\ \midrule
\multicolumn{1}{l|}{\multirow{2}{*}{\textcolor{red}{\ding{80}}\textbf{PanoSAM2}}}        & \multicolumn{1}{c|}{\cellcolor{tablehighlightgray}Hiera-T}      & \cellcolor{tablehighlightgray}64.3$\uparrow$\textcolor{blue}{4.9}          & \cellcolor{tablehighlightgray}59.2$\uparrow$\textcolor{blue}{5.3}          & \cellcolor{tablehighlightgray}69.3$\uparrow$\textcolor{blue}{4.4}  & 3.7       & 728.0  & 34.8        \\
\multicolumn{1}{l|}{}                                 & \multicolumn{1}{c|}{\cellcolor{tablehighlightgray}Hiera-S}      & \cellcolor{tablehighlightgray}\textbf{65.8}$\uparrow$\textcolor{blue}{5.6} & \cellcolor{tablehighlightgray}\textbf{59.9}$\uparrow$\textcolor{blue}{3.1} & \cellcolor{tablehighlightgray}\textbf{71.6}$\uparrow$\textcolor{blue}{8.0} & 4.3       & 788.3  & 29.2 \\  \bottomrule
\end{tabular}}
\end{table}

\section{Experiment}
\label{sec:experiment}

\subsection{Experimental Setup}
\noindent \textbf{Dataset.} We evaluate PanoSAM2 on two panoramic video object segmentation benchmarks: 360VOTS \cite{xu2025360vots} and PanoVOS \cite{yan2024panovos}. 360VOTS is a large-scale dataset that contains 290 sequences spanning 62 categories, totaling about 242K frames with an average duration of 28 seconds per sequence. It provides dense, pixel-wise ground-truth annotations. The official split assigns 170 sequences for training and 120 for testing. PanoVOS comprises 150 videos with about 14K frames and over 19K instance annotations from 35 categories, with an average duration of 20 seconds. The training set has 80 videos for training, and the validation set and test set each have 35 videos.

\noindent \textbf{Implementation Details.} PanoSAM2 is implemented in PyTorch \cite{paszke2019pytorch}. All components inherited from SAM2 \cite{ravi2024sam} are initialized from the released SAM2 training weights. We train with AdamW \cite{loshchilov2017decoupled} optimizer ($\beta$ = (0.9, 0.999), eps=1e-8, and the weight decay is 0.01) with an initial learning rate of 2e-4 and a StepLR scheduler, for a total of 80 epochs. To exercise long-horizon memory, we cap each sampled clip at 100 frames with a batch size of 4 and sample 400 clips per epoch. To warm up, in the first two epochs, the memory encoder is fed the ground-truth mask at every frame. In subsequent epochs, we correct the memory with a GT mask every 8 frames. After 20 epochs, we introduce LSMM for training. Our training is conducted on two NVIDIA A800-80GB GPUs and takes about 50 hours with the Hiera-T \cite{ryali2023hiera} backbone. We maintain a long-term memory size $L$ of 2. For geometric distortion augmentation, the $w_{max}$ and  $w_{min}$ of the pixel weight are set to 2.0 and 0.5, respectively. For optimization objective, $\lambda_{bce}$ is 0.5, $\lambda_{dice}$ is 0.5, 1.0 for $\lambda_{IoU}$, and 0.1 for $\lambda_{occ}$.

\begin{table}[tb]
\caption{\textbf{Comparisons between our method and existing approaches on PanoVOS validation and test datasets.} The best results are shown in \textbf{bold}. Evaluation metric subscript $s$ and $u$ denote scores in seen and unseen categories compared to the training dataset.}
\label{tab:panovos_result}
\centering
\resizebox{0.99\textwidth}{!}{
\begin{tabular}{lccccccccccc}
\toprule
\multicolumn{1}{l|}{\multirow{2}{*}{VOS Tracker}} & \multicolumn{1}{l|}{\multirow{2}{*}{Backbone}} & \multicolumn{5}{c|}{PanoVOS Validation}                                                            & \multicolumn{5}{c}{PanoVOS Test}                                              \\ \cmidrule(r){3-12} 
\multicolumn{1}{l|}{}                             & \multicolumn{1}{l|}{}                          & $\mathcal{J}\&\mathcal{F}$          & $\mathcal{J}_s$          & $\mathcal{F}_s$          & $\mathcal{J}_u$          & \multicolumn{1}{c|}{$\mathcal{F}_u$}          & $\mathcal{J}\&\mathcal{F}$          & $\mathcal{J}_s$          & $\mathcal{F}_s$         & $\mathcal{J}_u$          & $\mathcal{F}_u$         \\ \midrule
\multicolumn{12}{l}{\textit{Perspective-Video-Specified}}                                                                                                                                                                                                                               \\ \midrule
\multicolumn{1}{l|}{CFBI \cite{yang2020collaborative}}                         & \multicolumn{1}{c|}{ResNet-101}                & 35.8          & 34.6          & 44.8          & 24.2          & \multicolumn{1}{c|}{39.7}          & 19.1          & 18.2          & 26.1          & 12.2          & 19.8          \\
\multicolumn{1}{l|}{CFBI+ \cite{yang2020collaborative}}                        & \multicolumn{1}{c|}{ResNet-101}                & 41.3          & 38.0          & 47.9          & 32.5          & \multicolumn{1}{c|}{46.9}          & 30.9          & 30.8          & 42.7          & 21.4          & 28.5          \\
\multicolumn{1}{l|}{AFB-URR \cite{NEURIPS2020_liangVOS}}                      & \multicolumn{1}{c|}{ResNet-50}                 & 34.3          & 34.8          & 42.8          & 24.9          & \multicolumn{1}{c|}{34.5}          & 34.2          & 28.2          & 38.8          & 32.9          & 36.8          \\
\multicolumn{1}{l|}{STCN \cite{cheng2021rethinking}}                         & \multicolumn{1}{c|}{ResNet-50}                 & 52.0          & 51.2          & 60.8          & 41.5          & \multicolumn{1}{c|}{54.5}          & 50.8          & 43.6          & 56.5          & 49.3          & 53.7          \\
\multicolumn{1}{l|}{AOTT \cite{yang2021associating}}                         & \multicolumn{1}{c|}{MobileNet-V2}              & 65.6          & 59.4          & 68.3          & 59.7          & \multicolumn{1}{c|}{75.0}          & 53.4          & 49.3          & 61.6          & 47.5          & 55.1          \\
\multicolumn{1}{l|}{AOTS \cite{yang2021associating}}                         & \multicolumn{1}{c|}{MobileNet-V2}              & 67.7          & 61.2          & 70.0          & 62.4          & \multicolumn{1}{c|}{77.1}          & 55.9          & 53.2          & 65.1          & 48.6          & 57.0          \\
\multicolumn{1}{l|}{AOTB \cite{yang2021associating}}                         & \multicolumn{1}{c|}{MobileNet-V2}              & 67.6          & 62.3          & 72.0          & 61.5          & \multicolumn{1}{c|}{74.8}          & 55.4          & 53.5          & 64.2          & 47.7          & 56.0          \\
\multicolumn{1}{l|}{AOTL \cite{yang2021associating}}                         & \multicolumn{1}{c|}{MobileNet-V2}              & 66.6          & 61.4          & 71.1          & 59.4          & \multicolumn{1}{c|}{74.3}          & 53.8          & 50.0          & 60.3          & 47.8          & 57.1          \\
\multicolumn{1}{l|}{AOTL \cite{yang2021associating}}                         & \multicolumn{1}{c|}{Swin-Base}                 & 62.1          & 58.9          & 66.5          & 54.3          & \multicolumn{1}{c|}{68.8}          & 53.1          & 49.0          & 57.8          & 49.0          & 56.6          \\
\multicolumn{1}{l|}{AOTL \cite{yang2021associating}}                         & \multicolumn{1}{c|}{ResNet-50}                 & 65.3          & 61.9          & 71.4          & 56.4          & \multicolumn{1}{c|}{71.6}          & 54.6          & 52.9          & 63.2          & 47.5          & 54.9          \\
\multicolumn{1}{l|}{RDE \cite{li2022recurrent}}                          & \multicolumn{1}{c|}{ResNet-50}                 & 50.5          & 49.7          & 58.4          & 39.2          & \multicolumn{1}{c|}{54.9}          & 42.5          & 36.9          & 46.6          & 38.5          & 48.2          \\
\multicolumn{1}{l|}{XMem \cite{cheng2022xmem}}                         & \multicolumn{1}{c|}{ResNet-50}                 & 55.7          & 54.8          & 63.3          & 45.2          & \multicolumn{1}{c|}{59.7}          & 53.5          & 49.5          & 62.6          & 47.1          & 54.8          \\
\multicolumn{1}{l|}{SAM2 \cite{ravi2024sam}}                         & \multicolumn{1}{c|}{Hiera-T}                   & 74.2          & 64.5          & 79.4          & 68.0         & \multicolumn{1}{c|}{85.0}          & 70.6          & 62.3          & 79.8          & 63.8          & 76.5          \\
\multicolumn{1}{l|}{SAM2 \cite{ravi2024sam}}                         & \multicolumn{1}{c|}{Hiera-S}                   & 71.4          & 62.3          & 77.6          & 65.2          & \multicolumn{1}{c|}{80.3}          & 69.3          & 62.7          & 78.0          & 62.5          & 74.0          \\
\multicolumn{1}{l|}{SAM2Long \cite{ding2025sam2long}}                     & \multicolumn{1}{c|}{Hiera-T}                   & 74.4          & 63.7          & 78.6          & 69.3          & \multicolumn{1}{c|}{86.1}          & 70.9          & 62.8          & 79.7          & 64.2 & 76.9 \\
\multicolumn{1}{l|}{SAM2Long \cite{ding2025sam2long}}                     & \multicolumn{1}{c|}{Hiera-S}                   & 71.9          & 62.6          & 78.0          & \textbf{69.8}          & \multicolumn{1}{c|}{81.2}          & 69.7          & 63.1          & 78.6          & 62.8          & 74.3          \\ \midrule
\multicolumn{12}{l}{\textit{360-Video-Specified}}                                                                                                                                                                                                                                       \\ \midrule
\multicolumn{1}{l|}{PSCFormer \cite{yan2024panovos}-B}               & \multicolumn{1}{c|}{ResNet-50}                 & 74.0          & 66.4          & 80.4          & 66.2          & \multicolumn{1}{c|}{83.0}          & 56.8          & 49.4          & 62.7          & 52.4          & 62.5          \\
\multicolumn{1}{l|}{PSCFormer \cite{yan2024panovos}-L}              & \multicolumn{1}{c|}{ResNet-50}                 & 77.9          & 70.5          & 85.2          & 69.5 & \multicolumn{1}{c|}{86.4}          & 59.9          & 54.9          & 69.2          & 53.0          & 62.4          \\ \midrule
\multicolumn{1}{l|}{\multirow{2}{*}{\textcolor{red}{\ding{80}}\textbf{PanoSAM2}}}    & \multicolumn{1}{c|}{\cellcolor{tablehighlightgray}Hiera-T}                   & \cellcolor{tablehighlightgray}76.1$\uparrow$\textcolor{blue}{1.9}          & \cellcolor{tablehighlightgray}67.4 $\uparrow$\textcolor{blue}{2.9}         & \cellcolor{tablehighlightgray}83.4 $\uparrow$\textcolor{blue}{4.0}         & \cellcolor{tablehighlightgray}68.6$\uparrow$\textcolor{blue}{0.6}        & \multicolumn{1}{c|}{\cellcolor{tablehighlightgray}85.6$\uparrow$\textcolor{blue}{0.6}}          & \cellcolor{tablehighlightgray}71.9$\uparrow$\textcolor{blue}{1.3}           & \cellcolor{tablehighlightgray}64.8$\uparrow$\textcolor{blue}{2.5}           & \cellcolor{tablehighlightgray}80.5$\uparrow$\textcolor{blue}{0.7}           & \cellcolor{tablehighlightgray}64.3$\uparrow$\textcolor{blue}{0.5}         & \cellcolor{tablehighlightgray}\textbf{77.8} $\uparrow$\textcolor{blue}{1.3}         \\
\multicolumn{1}{l|}{}                             & \multicolumn{1}{c|}{\cellcolor{tablehighlightgray}Hiera-S}                   & \cellcolor{tablehighlightgray}\textbf{78.1}$\uparrow$\textcolor{blue}{6.7} & \cellcolor{tablehighlightgray}\textbf{71.2}$\uparrow$\textcolor{blue}{8.9} & \cellcolor{tablehighlightgray}\textbf{85.6}$\uparrow$\textcolor{blue}{8.0} & \cellcolor{tablehighlightgray}69.1$\uparrow$\textcolor{blue}{3.9}          & \multicolumn{1}{c|}{\cellcolor{tablehighlightgray}\textbf{86.7}$\uparrow$\textcolor{blue}{6.4}} & \cellcolor{tablehighlightgray}\textbf{73.4}$\uparrow$\textcolor{blue}{4.1} & \cellcolor{tablehighlightgray}\textbf{67.9}$\uparrow$\textcolor{blue}{5.2} & \cellcolor{tablehighlightgray}\textbf{84.0}$\uparrow$\textcolor{blue}{6.0} & \cellcolor{tablehighlightgray}\textbf{64.4}$\uparrow$\textcolor{blue}{1.9}          & \cellcolor{tablehighlightgray}77.4$\uparrow$\textcolor{blue}{3.4}          \\ \bottomrule
\end{tabular}}
\end{table}

\noindent \textbf{Evaluation Metrics.} We report $\mathcal{J}$, $\mathcal{F}$, and $\mathcal{J}\&\mathcal{F}$, following standard VOS metrics. $\mathcal{J}$ stands for $IoU$ between the predicted mask and the ground truth, measuring region overlap ratio. $\mathcal{F}$ is computed on mask boundaries, assessing contour accuracy. $\mathcal{J}\&\mathcal{F}$ averages $\mathcal{J}$ and $\mathcal{F}$, providing a composite score for overall segmentation quality.

\subsection{Experimental Results}
\noindent \textbf{Results on 360VOTS.} \cref{tab:360vots_result} compares the performance of PanoSAM2 with existing methods on the 360VOTS \cite{xu2025360vots} dataset, covering both VOS models for perspective videos and approaches tailored for 360-degree videos. The proposed PanoSAM2 demonstrates a clear and consistent improvement over previous methods, achieving state-of-the-art results across all evaluation metrics. Specifically, PanoSAM2 with the Hiera-S \cite{ryali2023hiera} backbone surpasses the baseline SAM2 by a significant margin, achieving a $\mathcal{J}\&\mathcal{F}$ score of 65.8, which represents an impressive improvement of 5.6 points. Also, in the 360-video-specified category, PanoSAM2 outperforms the PSCFormer method, whose $\mathcal{J}\&\mathcal{F}$ achieves 62.5, further highlighting its robustness and superiority in handling complex segmentation tasks.  Moreover, \cref{tab:360vots_result} summarizes the computational cost of several representative methods. PanoSAM2 builds upon the SAM2 \cite{ravi2024sam} framework with minimal architectural modification. Despite integrating panoramic-aware modules, the overall memory cost of PanoSAM2 remains close to its base SAM2 counterparts, increasing only modestly from 3.6G to 3.7G for Hiera-T and from 4.1G to 4.3G for Hiera-S. These results underline PanoSAM2’s substantial enhancement in 360 video segmentation and tracking, positioning it as a powerful and efficient solution for the 360VOS.

\begin{table}[tb]
\caption{Comparisons of\textbf{ generalization ability} between our method pretrained on 360VOTS and prior arts on PanoVOS validation and test dataset. The best results are shown in \textbf{bold}.}
\label{tab:zero_panovos_result}
\centering
\setlength{\tabcolsep}{1.8mm}
\resizebox{0.6\textwidth}{!}{
\begin{tabular}{l|ccc|ccc}
\toprule
\multirow{2}{*}{Method} & \multicolumn{3}{c|}{PanoVOS Validation}       & \multicolumn{3}{c}{PanoVOS Test}              \\ \cmidrule(r){2-7} 
                        & $\mathcal{J}\&\mathcal{F}$          & $\mathcal{J}$             & $\mathcal{F}$             & $\mathcal{J}\&\mathcal{F}$          & $\mathcal{J}$             & $\mathcal{F}$             \\ \midrule
PerSAM \cite{zhang2023personalize}                  & 19.1          & 14.9          & 23.5          & 19.5          & 15.6          & 23.3          \\
SAM-PT \cite{rajivc2025segment}                  & 47.5          & 41.4          & 53.7          & 41.0          & 35.7          & 46.4          \\
SAM-I2V  \cite{mei2025sam}               & 48.9          & 42.8          & 55.0          & 40.5          & 34.9          & 46.1          \\
SAM2  \cite{ravi2024sam}               & 70.6          & 63.4          & 77.8          & 66.3          & 59.4          & 73.2          \\
XMem  \cite{cheng2022xmem}               & 52.1          & 46.9          & 57.2          & 44.4          & 39.6          & 49.1          \\
PSCFormer  \cite{yan2024panovos}               & 69.5          & 62.4          & 76.5          & 64.1          & 58.3          & 69.9          \\
\textcolor{red}{\ding{80}}\textbf{PanoSAM2}                & \cellcolor{tablehighlightgray}\textbf{71.3} & \cellcolor{tablehighlightgray}\textbf{63.0} & \cellcolor{tablehighlightgray}\textbf{79.5} & \cellcolor{tablehighlightgray}\textbf{67.9} & \cellcolor{tablehighlightgray}\textbf{61.6} & \cellcolor{tablehighlightgray}\textbf{74.2} \\ \bottomrule
\end{tabular}}
\end{table}

\begin{figure}[t!]
    \centering
    \includegraphics[width=\textwidth]{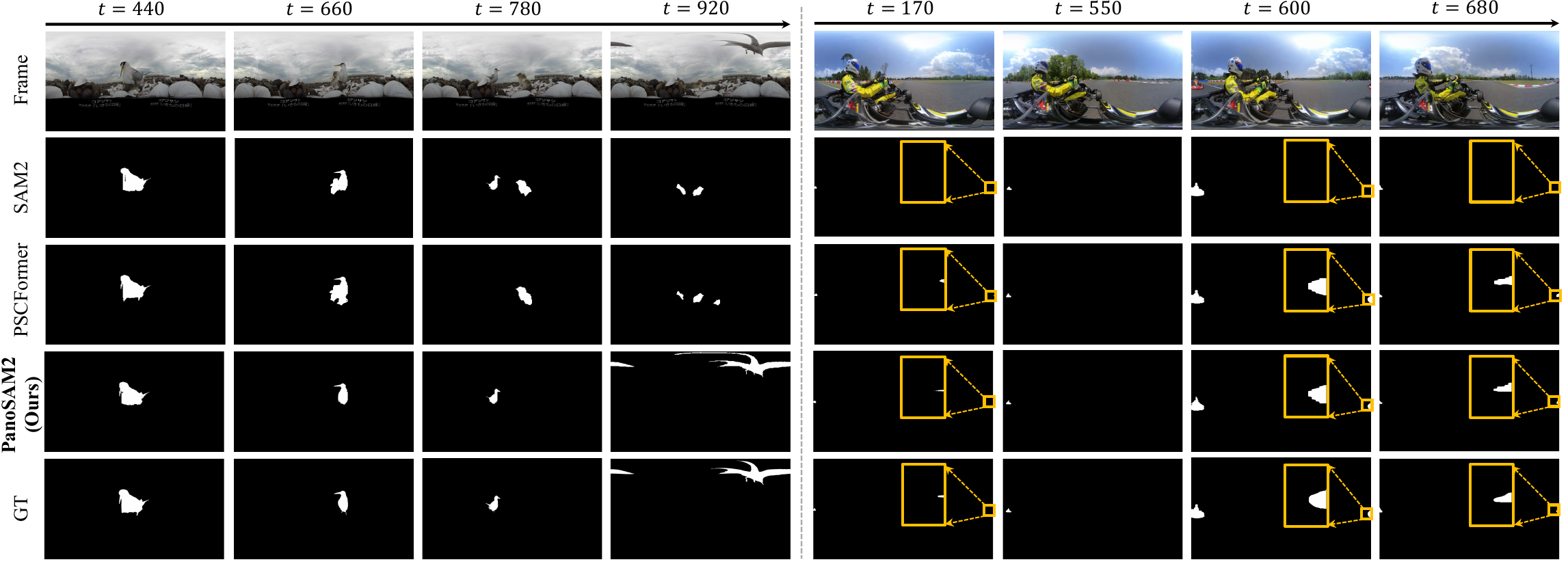}
    \caption{\textbf{Qualitative comparison between PanoSAM2 and other VOS models.} The orange boxes highlight the areas where the target object straddles the 0°/360° seam, showing PanoSAM2's improved segmentation precision in complex scenes at different time steps.}
    \label{fig:visualization_comparsion}
    \vspace{-5pt}
\end{figure}

\noindent \textbf{Results on PanoVOS.} \cref{tab:panovos_result} presents a comparison between PanoSAM2 and existing methods on the PanoVOS validation and test datasets. PanoSAM2 demonstrates substantial improvements over prior techniques, achieving state-of-the-art performance across multiple evaluation metrics. On the PanoVOS validation dataset, PanoSAM2 (Hiera-S) achieves a $\mathcal{J}\&\mathcal{F}$ score of 78.1, surpassing the SAM2 baseline by 6.7 points and indicating stronger spatial stability. In the PanoVOS test dataset, PanoSAM2 (Hiera-S) achieves an outstanding $\mathcal{J}\&\mathcal{F}$ score of 73.4, improving by 4.1 points over SAM2. Furthermore, PanoSAM2 also excels SAM2 in the unseen category tracking in terms of $\mathcal{J}_u$ and $\mathcal{F}_u$. When compared with other 360VOS methods, PanoSAM2 also outperforms PSCFormer. PanoSAM2’s result is an improvement over PSCFormer-Base and PSCFormer-Large in most evaluation metrics. \textit{Similar to the findings in the SAM2 paper, our experiments show that SAM2 (Hiera-T) outperforms SAM2 (Hiera-S), which is consistent with the observation that smaller models can sometimes outperform larger ones, as shown in experiments of SAM2’s paper.} These results confirm the effectiveness of PanoSAM2 in handling 360VOS tasks. The performance gains across multiple metrics demonstrate that PanoSAM2 sets new benchmarks for 360VOS, offering a significant advancement over current methods in the 360VOS challenge.

% \begin{figure}[t!]
%     \
%     \includegraphics[width=0.5\linewidth]{open_world_visualization.pdf}
%     \caption{Visual results on self-captured \textbf{open-world} 360 scene.}
%     \label{fig:open_world_visualization}
% \end{figure}

\begin{figure*}[t!]
    \centering
    \includegraphics[width=\textwidth]{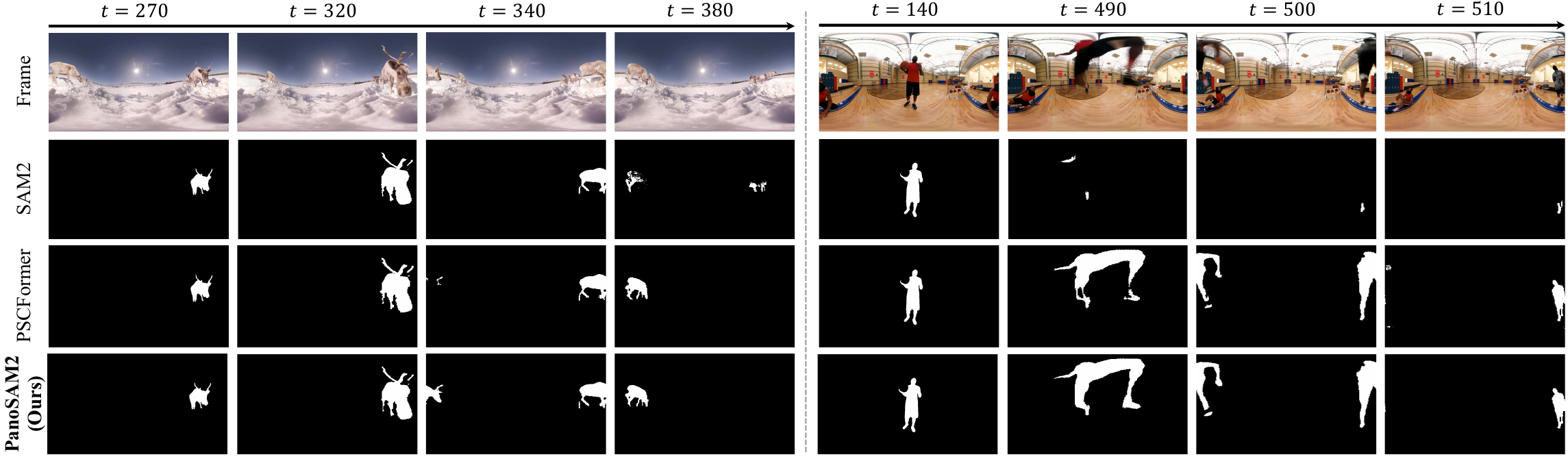}
    \caption{Visualization of \textbf{zero-shot results} from PanoSAM2 and other VOS models on the PanoVOS dataset.}
    \label{fig:zero_shot_visualization}
    \vspace{-15pt}
\end{figure*}

\begin{figure}[t]
\centering
\begin{minipage}[t]{0.49\linewidth}
\centering
\footnotesize
\includegraphics[width=\linewidth]{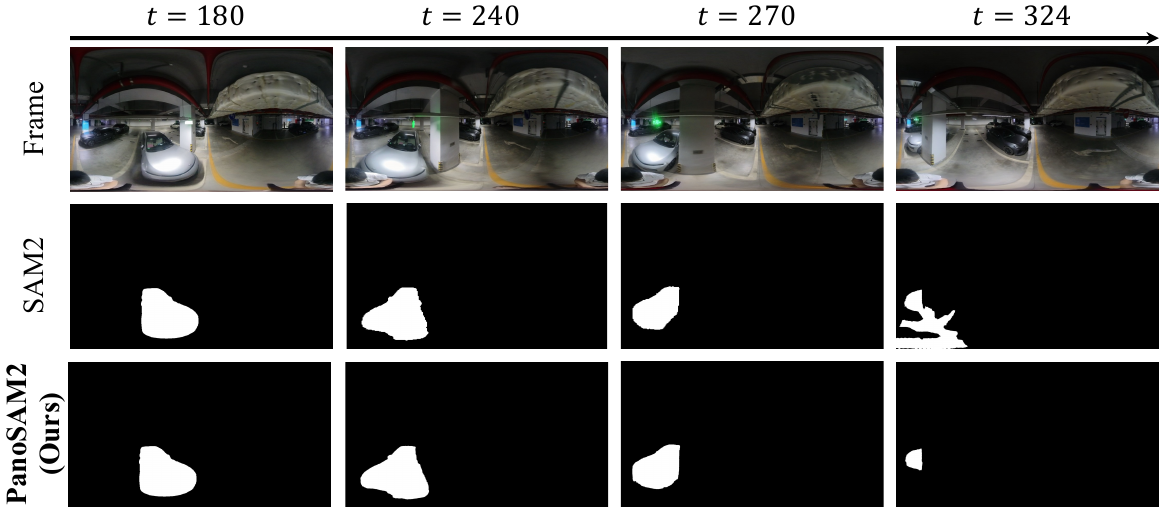}
\captionof{figure}{Visual results on self-captured \textbf{open-world} 360 scene.}
\label{fig:open_world_visualization}
\end{minipage}
\hfill
\begin{minipage}[t]{0.48\linewidth}
\centering
\footnotesize
\includegraphics[width=\linewidth]{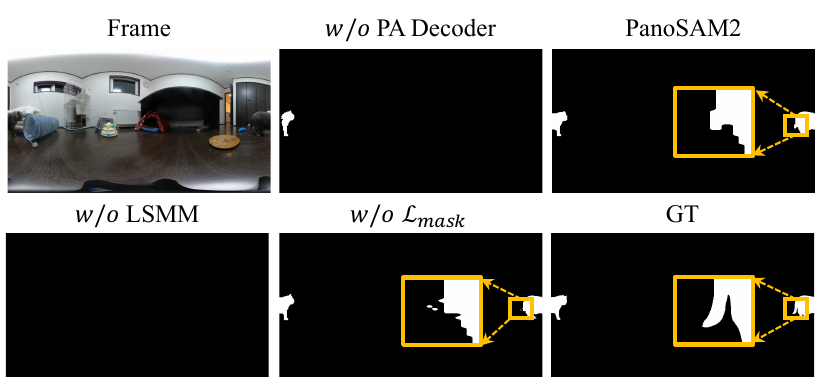}
\captionof{figure}{Ablation visualizations of proposed modules.}
\label{fig:ablation_visualization}
\end{minipage}
\vspace{-15pt}
\end{figure}

\noindent \textbf{Qualitative Comparison.} As visualized in \cref{fig:visualization_comparsion}, PanoSAM2 shows a clear advantage on 360VOS task. Unlike SAM2 and PSCFormer, it isolates the target without distraction from similar objects. PanoSAM2 also maintains accurate segmentation even in complex scenarios when only a small part of the object crosses the boundary. This is evident in the highlighted regions, where it successfully tracks and segments the intended target, demonstrating robustness to the unique challenges of panoramic images.

\noindent \textbf{Zero-Shot Comparison.} \cref{tab:zero_panovos_result} compares the generalization ability of our PanoSAM2 pretrained on the 360VOTS training dataset with other VOS models on the PanoVOS validation and test datasets. Our model consistently demonstrates superior performance across evaluation metrics, clearly showing that PanoSAM2 possesses strong zero-shot capability and robust transferability, as further visualized in \cref{fig:zero_shot_visualization}. We also show the result of PanoSAM2 on the open-world 360 video taken by ourselves in \cref{fig:open_world_visualization}.

\subsection{Ablation Studies}
We conduct a series of ablations on the 360VOTS test set using the Hiera-T backbone. We vary one factor at a time to assess the contributions of the PA decoder, LSMM, and the distortion-aware mask loss $\mathcal{L}_{\text{mask}}$. We also examine model sensitivity to hyperparameters.

\begin{table}[t!]
\caption{\textbf{Ablation study on key components of PanoSAM2.} The best result is shown in \textbf{bold}.}
\label{tab:ablation_result}
\centering
\resizebox{0.5\textwidth}{!}{
\begin{tabular}{ccc|ccc}
\toprule
PA Decoder & LSMM & $\mathcal{L}_{mask}$ & $\mathcal{J}\&\mathcal{F}$ & $\mathcal{J}$    & $\mathcal{F}$    \\ \midrule
           &      &           & 59.4 & 53.9 & 64.9 \\
\checkmark        &      &           & 62.7 & 57.6 & 67.8 \\
           & \checkmark  &           & 61.6 & 57.4 & 65.8 \\
           &      & \checkmark       & 60.1 & 54.7 & 65.5 \\
\checkmark        & \checkmark  & \checkmark       & \cellcolor{tablehighlightgray}\textbf{64.3} & \cellcolor{tablehighlightgray}\textbf{59.2} & \cellcolor{tablehighlightgray}\textbf{69.3} \\ \bottomrule
\end{tabular}}
\end{table}

\begin{table}[t]
\begin{minipage}[t]{0.46\linewidth}
\vspace{-1pt}
\caption{\textbf{Ablation study on innovative elements of PA Decoder}. The best result is shown in \textbf{bold}.}
\footnotesize
\label{tab:ablation_pa}
\centering
\resizebox{0.8\textwidth}{!}{
\begin{tabular}{cc|ccc}
\toprule
PC Block & $\mathcal{Y}^{T-1}_{pano}$    & $\mathcal{J}\&\mathcal{F}$                    & $\mathcal{J}$                        & $\mathcal{F}$                         \\\midrule
         &      & 62.2                     & 56.3                     & 68.1                     \\
\checkmark     &      & 63.8 & 58.7 & 68.9 \\
         & \checkmark & 62.9 & 57.4 & 68.6 \\
\checkmark     & \checkmark  & \cellcolor{tablehighlightgray}\textbf{64.3} & \cellcolor{tablehighlightgray}\textbf{59.2} & \cellcolor{tablehighlightgray}\textbf{69.3}                     \\ \bottomrule
\end{tabular}}
\end{minipage}
\hfill
\begin{minipage}[t]{0.52\linewidth}
\caption{\textbf{Impact of using different long-term memory sizes in LSMM.} The best result is shown in \textbf{bold}.}
\footnotesize
\label{tab:ablation_lsmm}
\centering
\resizebox{0.9\textwidth}{!}{
\begin{tabular}{c|ccc}
\toprule
Long-Term Memory Size $L$ & $\mathcal{J}\&\mathcal{F}$                    & $\mathcal{J}$                        & $\mathcal{F}$              \\ \midrule
$w/o$                       & 63.4          & 58.8          & 68.0          \\ 
1                       & 64.0          & 59.1          & 68.9          \\ 
2                       & \cellcolor{tablehighlightgray}\textbf{64.3} & \cellcolor{tablehighlightgray}\textbf{59.2} & \cellcolor{tablehighlightgray}\textbf{69.3} \\
3                       & 63.7          & 59.1          & 68.3          \\ \bottomrule
\end{tabular}}
\end{minipage}
\end{table}

% \begin{figure}
%   \centering
%   \includegraphics[width=0.5\linewidth]{ablation_visualization.pdf}
%    \caption{Ablation visualizations of proposed modules.}
%    \label{fig:ablation_visualization}
% \end{figure}

\noindent \textbf{Effectiveness of Key Components.} 
\cref{tab:ablation_result} reports performance gains of our components and \cref{fig:ablation_visualization} visualizes the effect of these ablation studies. Starting from the SAM2 baseline, the PA decoder provides the largest single boost, adding \textbf{+3.3} in terms of $\mathcal{J}\&\mathcal{F}$, and keeps the tracking seam-consistent. LSMM alone yields a modest \textbf{+2.2} and prevents object drift. Replacing the naive cross-entropy mask loss with the distortion-aware $\mathcal{L}_{mask}$ lifts $\mathcal{J}\&\mathcal{F}$ from 59.4 to 60.1 (\textbf{+0.7}) and produces a mask with more precise boundary. Enabling all three components yields a total \textbf{+4.9}. All other settings are held fixed to isolate effects. These findings highlight specific roles of different components in improving tracker performance.

\noindent \textbf{Influence of PA Decoder Elements.} 
% As shown in \cref{tab:ablation_pa}, the incorporation of PC Block alone improves $\mathcal{J}\&\mathcal{F}$ from 62.2 to 63.8 (\textbf{+1.6}), while the incorporation of the mask of the last time step $\mathcal{Y}^{T-1}_{pano}$ alone raises it to 62.9 (\textbf{+0.7}). When combined, performance increases to 64.3 (\textbf{+2.1}), demonstrating the strong complementarity of the two components. This clearly indicates that the PC Block effectively enhances seam-consistent feature aggregation, whereas $\mathcal{Y}^{T-1}_{pano}$ provides valuable temporal and distortion-aware guidance, jointly leading to more coherent panoramic understanding and smoother object boundaries across consecutive frames.
As shown in \cref{tab:ablation_pa}, the incorporation of PC Block alone improves $\mathcal{J}\&\mathcal{F}$ from 62.2 to 63.8 (\textbf{+1.6}), while the incorporation of the mask of the last time step $\mathcal{Y}^{T-1}{pano}$ alone raises it to 62.9 (\textbf{+0.7}). When combined, performance increases to 64.3 (\textbf{+2.1}), demonstrating their complementarity. This indicates that the PC Block enhances seam-consistent feature aggregation, while $\mathcal{Y}^{T-1}{pano}$ provides temporal guidance, jointly improving panoramic understanding and boundary consistency.

\noindent \textbf{Impact of Long-Term Memory Size $L$.}  
\cref{tab:ablation_lsmm} indicates that introducing long-term memory notably enhances performance, confirming its importance for stable temporal reasoning. Compared with no LSMM, the best configuration at $L=2$ improves $\mathcal{J}\&\mathcal{F}$ from 63.4 to \textbf{64.3} (\textbf{+0.9}), with $\mathcal{J}$ and $\mathcal{F}$ also rising to \textbf{59.2} and \textbf{69.3}, respectively. However, when $L$ grows larger ($L=3$), the gain diminishes slightly, suggesting that excessive long-term memory may dilute short-term information critical for precise memory conditioning.

\subsection{Failure Cases}

\begin{figure}[t]
    \centering
    \includegraphics[width=\textwidth]{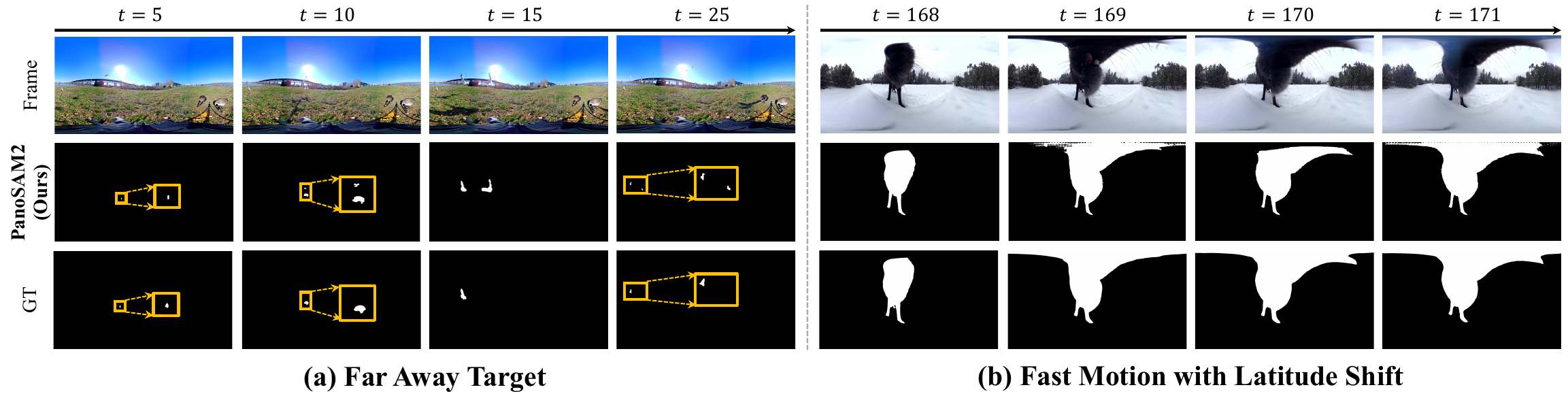}
    \caption{\textbf{Examples of failure cases.} Orange bounding boxes highlight and zoom in on the small mask region.}
    \label{fig:supply_failure_case}
    \vspace{-10pt}
\end{figure}

\noindent \textbf{Far Away Object.} The first example of \cref{fig:supply_failure_case} demonstrates a particularly challenging scenario in which distant objects with highly similar appearances interact or partially occlude one another. When the target pigeon and another overlap or move close together, PanoSAM2 occasionally misassigns masks or swaps identities, as their visual cues provide limited discriminative information.

\noindent \textbf{Fast Motion with Latitude Shift.} 
 As illustrated in the second example of \cref{fig:supply_failure_case}, when the animal rapidly moves upward toward the camera, its position shifts abruptly from low to high latitudes, causing severe projection distortion. The previous-frame mask becomes heavily stretched and no longer serves as a reliable reference, resulting in fragmented predictions of PanoSAM2.

\section{Conclusion and Future Work}
We introduced PanoSAM2, a novel 360VOS framework based on our distortion- and memory-aware adaptation strategies of SAM2 to achieve reliable 360VOS while retaining SAM2’s user-friendly prompting design. 
Extensive experiments demonstrated that PanoSAM2 delivered robust and consistent performance across diverse 360 datasets.
Overall, this work underscored the importance of jointly modeling geometric distortion, temporal coherence, and memory dynamics when adapting foundation VOS models to omnidirectional perception, paving the way toward more reliable and generalizable panoramic video understanding.

\noindent \textbf{Future Work.} 
We will extend PanoSAM2 toward multi-object tracking and diverse prompt understanding. While proposed PanoSAM2 handles single-object segmentation effectively, reasoning over multiple interacting targets remains a challenging yet valuable direction for real-world scenes. Incorporating richer prompt types -- such as points, boxes, or scribbles -- could enable more flexible interaction, enhancing adaptability across tasks and environments. 

% \section*{Acknowledgements}
% Please insert your acknowledgments here.

\bibliographystyle{splncs04}
\bibliography{main}

\newpage
\appendix

\section{More Details of Methodology}
\label{sec:supply_methodology}
Due to space limitations in the main paper, we provide additional explanations of the novel design within the PanoSAM2 framework using pseudocode. \cref{subsec:supply_lsmm} provides further details about the Long-Short Memory Module (LSMM), while \cref{subsec:supply_loss} elaborates on the design aspects of Distortion-Guided Mask Loss.

\subsection{More Details of LSMM}
\label{subsec:supply_lsmm}
This section provides some details about the LSMM module, including its underlying motivation and the design of our Occlusion Sample Strategy, which jointly enhance long-term awareness in challenging 360 VOS scenarios.

\noindent \textbf{Motivation.} Many existing approaches adopt long-term memory mechanisms \cite{cheng2024putting, cheng2022xmem, bekuzarov2023xmem++, yan2024panovos, li2022recurrent} that periodically store frames into a memory bank, but this strategy leads to a substantial increase in computational and storage costs. In contrast, SAM2 \cite{ravi2024sam} retains only the first frame and the most recent six frames, which significantly limits its ability to capture long-range temporal information. Although SAM2Long \cite{ding2025sam2long} introduces tracking branches derived from multi-mask outputs to correct error accumulation, it incurs considerable overhead in both speed and memory usage. To address these limitations, we propose LSMM, which enriches short-term memory with distilled long-term cues, stabilizing identity while preserving responsiveness to new inputs, thereby improving temporal coherence in challenging panoramic scenes.

\begin{algorithm}[t]
\DontPrintSemicolon
\KwInput{NonCondOutputs marks dictionary of past frame outputs except those in short-term memory and the first prompted frame, and $L$ denotes the long-term memory size. 
}
\KwOutput{SelectedFrameIdx marks the list of selected memory indices, and SelectedObjPtr marks the list of selected object pointers.
}
LongCand $\leftarrow$ []

Scores   $\leftarrow$ []

SelectedFrameIdx $\leftarrow$ []

SelectedObjPtr $\leftarrow$ []

\ForEach{(frame\_idx, out) $\in$ NonCondOutputs}{
    LongCand.append((frame\_idx, out.obj\_ptr))
    
    Scores.append(exp(out.obj\_score))
}
\If{LongCand is empty or $L = 0$}{
    return None, None
}
\ForEach{$k$ $\in$ $1\dots L$}{
    $W$ $\leftarrow$ sum(Scores)
    
    $r$ $\leftarrow$ UniformSample(0, $W$)
    
    $c$ $\leftarrow$ 0
    
    \ForEach{$i$ $\in$ $\{1,\dots \left| \text{LongCand} \}\right|\}$}{
        $c$ $\leftarrow$ $c$ $+$ Scores[i]
        
        \If{$c$ $\geq$ $r$}{
            SelectedFrameIdx.append(LongCand[i][0])
            
            SelectedObjPtr.append(LongCand[i][1])
            
            break
        }
    }
}
\Return SelectedFrameIdx and SelectedObjPtr
\caption{Occlusion Sample Strategy}
\label{alg:occ_sample}
\end{algorithm}

\noindent \textbf{Occlusion Sample Strategy.} The proposed Occlusion Sample Strategy aims to identify long-term key frames by performing weighted sampling based on predicted occlusion scores. For each historical frame, the model predicts an occlusion score $\mathcal{O}^{T}$ ($T$ for time step) that reflects the likelihood of the target object being visible. A higher score corresponds to a stronger and more reliable object presence, indicating that the frame contains salient object information suitable for enriching long-term memory. Building on this idea, the strategy first collects all candidate frames before the current step and computes their sampling weights by exponentiating the occlusion scores to enhance the contrast between confident and uncertain frames. As illustrated in \cref{alg:occ_sample}, the algorithm maintains two lists: one storing frame–object pointer pairs and the other storing the corresponding weights. It then performs iterative weighted sampling without replacement: at each iteration, the cumulative sum of weights is computed, a random value is drawn within this range, and the frame whose cumulative weight first exceeds the random value is selected as a long-term memory slot. After selection, the chosen frame and its weight are removed to avoid reselection, and the process repeats until the required long-term memory size is reached. This strategy effectively injects high-confidence, occlusion-robust cues into the long-term memory while keeping computational overhead low.

\begin{algorithm}[t]
\DontPrintSemicolon
\KwInput{Panoramic GT mask $M$; weight bounds $(w_{\min}, w_{\max})$; decay power $\alpha$.}
\KwOutput{Distortion-guided weight map $W$.}

$M^{\text{sr}} \leftarrow \tau(M)$, $\tau$ is the BFoV projection\;
Initialize $W^{\text{sr}}$ with zeros, same size as $M^{\text{sr}}$\;

$\text{FG} \leftarrow \{p \mid M^{\text{sr}}(p) = 1\}$, \quad
$\text{BG} \leftarrow \{p \mid M^{\text{sr}}(p) = 0\}$\;

$w_f \leftarrow \dfrac{|\text{BG}|}{|\text{FG}|}$\;
$w_f \leftarrow \text{clip}(w_f, w_{\min}, w_{\max})$\;
$w_b \leftarrow \dfrac{1}{w_f}$\;

\ForEach{$p \in \text{FG}$}{
    $W^{\text{sr}}(p) \leftarrow w_f$\;
}

$D \leftarrow \text{DistanceTransform}(\text{BG})$

\eIf{$\max(D) > 0$}{
    $D_{\text{norm}} \leftarrow \frac{D}{\max(D)}$\;
}{
    $D_{\text{norm}} \leftarrow D$\;
}

$W_{\text{bg}} \leftarrow w_b + (w_f - w_b)(1 - D_{\text{norm}})^{\alpha}$\;

\ForEach{$p \in \text{BG}$}{
    $W^{\text{sr}}(p) \leftarrow W_{\text{bg}}(p)$\;
}

$W \leftarrow \tau^{-1}(M^{\text{sr}}, W^{\text{sr}})$\;

Replace zeros in $W$ with $\min(W^{\text{sr}})$\;

\Return $W$\;

\caption{Distortion-Guided Pixel Weight Generation}
\label{alg:distortion_weight}
\end{algorithm}

\subsection{More Details of Distortion-Guided Mask Loss}
\label{subsec:supply_loss}
This section provides more details about Distortion-Guided Mask Loss, including our motivation and the detailed construction of pixel weight $W$. It aligns the tracking optimization objective with spherical sampling of ERP.

\noindent \textbf{Motivation.} Existing panoramic mask losses \cite{yan2024panovos, zhang2024goodsam} simply mimic perspective-based formulations \cite{lu2020video, oh2019video, ravi2024sam} and overlook distortion-aware characteristics, leaving important spatial bias unmodeled and limiting the accuracy of segmentation in equirectangular images.

\noindent \textbf{Distortion-Guided Pixel Weight Generation.} \cref{alg:distortion_weight} assigns pixel-wise weight $W$ to the ground truth panoramic mask $\mathcal{Y}^{T}_{gt}$ by incorporating both object balance and distortion awareness. The algorithm first extracts a search-region mask that better matches local panoramic geometry. It then separates foreground and background pixels and computes an adaptive foreground weight based on their ratio, ensuring that objects of different sizes contribute fairly to the loss. Foreground pixels are directly assigned this weight. For background pixels, a distance transform is applied to measure how far each pixel is from the object boundary. Using this distance, the algorithm gradually adjusts background weights: pixels near the boundary receive higher importance, while distant pixels receive lower weights, controlled by a decay factor. The generated weight map is then reprojected back to the panoramic space, and zero values are replaced with the minimum weight to maintain stability. This process produces a balanced and distortion-sensitive weight map that improves training for panoramic tracking.

\section{More Details of Experiments}
\label{sec:supply_experiment}
Due to space limitations in the main paper, this section provides additional extended hyperparameter evaluations (\cref{subsec:suppl_hyperparameter_settings}), further experiments related to the bounding field-of-view (BFoV) \cite{xu2025360vots} (\cref{subsec:suppl_bfov_extensive}), and experiments about the performance on perspective videos (\cref{subsec:perspective_video}).

\subsection{More Details of Hyperparameter Settings}
\label{subsec:suppl_hyperparameter_settings}

\begin{table}[t]
\begin{minipage}[t]{0.45\textwidth}
\centering
\caption{Influence of different distortion-guided loss settings on 360VOTS test dataset.}
\label{tab:supply_distortion_settings}
\begin{tabular}{cc|ccc}
\toprule
$w_{min}$ & $w_{max}$ & $\mathcal{J}\&\mathcal{F}$                    & $\mathcal{J}$                        & $\mathcal{F}$          \\ \midrule
1.00 & 1.00              & 63.8          & 59.1          & 68.5          \\
0.80 & 1.25              & 63.9          & 59.1          & 68.7          \\
0.50 & 2.00               & \cellcolor{tablehighlightgray}\textbf{64.3} & \cellcolor{tablehighlightgray}\textbf{59.2} & \cellcolor{tablehighlightgray}\textbf{69.3} \\
0.25 & 4.00              & 63.6          & 58.8          & 68.4          \\ \bottomrule
\end{tabular}
\end{minipage}
\hfill
\begin{minipage}[t]{0.46\textwidth}
\caption{Influence of different $\lambda_{occ}$ settings on 360VOTS test dataset. \\}
\label{tab:supply_occlusion_weight}
\centering
\begin{tabular}{c|ccc}
\toprule
$\lambda_{occ}$ & $\mathcal{J}\&\mathcal{F}$ & $\mathcal{J}$    & $\mathcal{F}$    \\ \midrule
0.0      & 63.1 & 58.5 & 67.7 \\
0.1      & \textbf{64.3} & 59.2 & 69.3 \\
0.5      & 64.2 & 58.9 & \textbf{69.5} \\ 
1.0      & 64.2 & \textbf{59.3} & 69.1 \\ \bottomrule
\end{tabular}
\end{minipage}
\end{table}

\begin{table}[t]
\centering
\caption{\footnotesize $\mathcal{J}\&\mathcal{F}$ of SAM2-based tackers on perspective VOS datasets, where $^\text{‡}$ stands for fine-tuned on 360VOTS.}
\label{tab:2d_video}
\resizebox{0.6\linewidth}{!}{
\begin{tabular}{l|c|ccc}
\toprule
Model & Backbone & LVOSv2 val & SA-V val & SA-V test \\ \midrule
SAM2  & Hiera-T     & \textbf{77.3}     & \textbf{75.2}     & \textbf{76.5}      \\ \midrule
SAM2$^\text{‡}$  & Hiera-T      & 70.1        & 64.5     & 65.4      \\ \midrule
PanoSAM2$^\text{‡}$ & Hiera-T   & 68.7        & 63.5     & 64.1      \\ \bottomrule
\end{tabular}}
\end{table}

\noindent \textbf{Influence of Distortion-Guided Loss Settings.}
As shown in \cref{tab:supply_distortion_settings}, adjusting the pixel-weight range in the distortion-guided loss has a clear and consistent impact on segmentation performance. The optimal configuration at $(w_{min}=0.5, w_{max}=2.0)$ raises $\mathcal{J}\&\mathcal{F}$ from 63.8 to 64.3 (\textbf{+0.5}), and simultaneously achieves the highest $\mathcal{J}$ and $\mathcal{F}$ scores of \textbf{59.2} and \textbf{69.3}. This demonstrates that a moderate weighting interval provides a balanced emphasis on distorted regions, strengthening both boundary quality and foreground consistency without introducing instability. In contrast, when the weighting range becomes overly wide (e.g., $w_{max}=4.0$), the model tends to over-prioritize high-distortion foreground areas, reducing the relative contribution of background cues. This imbalance weakens global supervision and leads to a slight drop in overall accuracy, highlighting the importance of carefully choosing the weighting bounds.

\noindent \textbf{Impact of Different $\lambda_{occ}$.} 
The choices of $\lambda_{bce}$, $\lambda_{dice}$, and $\lambda_{iou}$ is directly followed SAM2 \cite{ravi2024sam}. The occlusion loss weight $\lambda_{occ}$, as shown in \cref{tab:supply_occlusion_weight}, directly affects the tracking result. Since LSMM relies on reliable occlusion scores to select long-term frames, when the weight is removed entirely ($\lambda_{occ}=0$), the model suffers a clear drop in performance, with $\mathcal{J}\&\mathcal{F}$ decreasing to 63.1, indicating that the occlusion-aware supervision is essential for guiding memory selection. In contrast, when $\lambda_{occ}$ is set to 0.1, 0.5, or 1.0, the overall performance remains relatively consistent. These settings all produce strong results, with each achieving either the best $\mathcal{J}$ or best $\mathcal{F}$ score. Among them, $\lambda_{occ}=0.1$ yields the highest combined $\mathcal{J}\&\mathcal{F}$ of 64.3, which we adopt as our default configuration. This suggests that a small but non-zero occlusion weight is sufficient to provide meaningful guidance while avoiding overly strong regularization.

\subsection{Extensive Experiment about Perspective Videos}
\label{subsec:perspective_video}
As shown in \cref{tab:2d_video}, there is a performance drop on standard perspective videos, with models like SAM2 and PanoSAM2 seeing significant reductions in $\mathcal{J}\&\mathcal{F}$ scores across the LVOSv2 and SA-V validation and test sets. This suggests some degree of forgetting when transferring to perspective settings, which may impact the model’s generalization. However, since our primary focus is 360VOS, this trade-off is acceptable, and the performance on the intended task, particularly on the 360VOS dataset.

\subsection{Extensive Experiment about BFoV}
\label{subsec:suppl_bfov_extensive}
The bounding field-of-view (BFoV) strategy \cite{xu2025360vots} provides a handcrafted mechanism that selects the next search region based on the prediction of the previous frame, allowing conventional 2D VOS models to be applied directly to omnidirectional scenarios. When combined with BFoV, as shown in \cref{tab:supply_bfov_comparison}, perspective-based trackers achieve noticeable improvements in $\mathcal{J}\&\mathcal{F}$ scores. However, the running speed of these models is significantly reduced, with FPS values clearly dropping as shown in the table. SAM2+BFoV consistently outperforms these methods by a large margin, achieving the best $\mathcal{J}\&\mathcal{F}$ score of \textbf{73.6} with the Hiera-S backbone. This highlights SAM2’s strong generalization ability and its compatibility with panoramic inputs, even without explicit 360-degree design. \textbf{Despite these improvements, BFoV introduces several drawbacks, such as reduced inference speed due to repeated cropping and projection. Moreover, its performance degrades when the target undergoes severe occlusion, as the restricted view may entirely exclude the object. To address these challenges, we propose a novel integration of BFoV into the loss design of PanoSAM2, mitigating both the computational overhead and occlusion-related limitations.}

\section{More Discussions}
\label{sec:supply_discussion}

\begin{table}[!t]
\centering
\begin{minipage}[t]{0.48\textwidth}
\vspace{0pt}
\centering
\caption{Comparision of different VOS models with BFoV framework \cite{xu2025360vots} on 360VOTS test dataset, where $^\text{‡}$ stands for fine-tuned.}
\label{tab:supply_bfov_comparison}
\centering
\resizebox{\linewidth}{!}{
\begin{tabular}{l|c|ccc|c}
\toprule
VOS Tracker & Backbone                   & $\mathcal{J}\&\mathcal{F}\uparrow$           & $\mathcal{J}\uparrow$             & $\mathcal{F}\uparrow$              & FPS$\downarrow$  \\ \midrule
STCN$^\text{‡}$\cite{cheng2021rethinking}       & \multirow{2}{*}{ResNet-50} & 60.9          & 55.0          & 66.7          & 23.8 \\
STCN+BFoV   &                            & 64.0          & 58.4          & 69.6          & 9.6  \\ \midrule
XMem$^\text{‡}$\cite{cheng2022xmem}       & \multirow{2}{*}{ResNet-50} & 65.0          & 59.6          & 70.3          & 22.5 \\
XMem+BFoV   &                            & 72.6 & 66.5          & 78.6 & 5.8  \\ \midrule
SAM2$^\text{‡}$\cite{ravi2024sam}       & \multirow{2}{*}{Hiera-S}   & 60.2          & 56.8          & 63.6          & 39.3 \\
SAM2+BFoV   &                            & \textbf{73.6}          & \textbf{69.4} & \textbf{78.9}          & 5.5  \\ \midrule
PanoSAM2$^\text{‡}$   & Hiera-S                    & 65.8          & 59.9          & 71.6                               & 29.2 \\ \bottomrule
\end{tabular}}
\vspace{-10pt}
\end{minipage}
\hfill
\begin{minipage}[t]{0.51\textwidth}
\vspace{0pt}
\centering
\includegraphics[width=\linewidth]{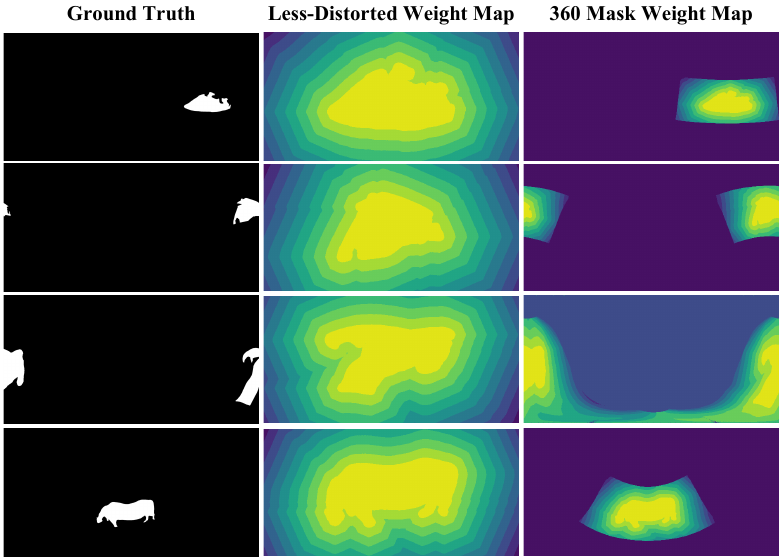}
{\captionof{figure}{Visualizations of distortion-guided 360 mask weight heatmap.}
\label{fig:supply_360_mask_weight}}
\end{minipage}
\vspace{-10pt}
\end{table}

\subsection{More Visualizations of 360 Mask Weight}
\cref{fig:supply_360_mask_weight} provides additional visualizations of our distortion-guided 360 mask weight maps. Unlike conventional mask weighting, which treats all pixels uniformly, our approach explicitly accounts for projection-induced distortion and the uneven spatial distribution of foreground and background regions in panoramic imagery. As illustrated in the figure, the less-distorted weight maps produced in the local search region emphasize object boundaries and nearby background pixels through smoothly decaying weights, while the reprojected 360 weight maps further highlight regions affected by panoramic stretching. This allows the model to allocate stronger supervision to areas where geometric distortion is more severe or where object structure is harder to preserve. Overall, these visualizations demonstrate that the distortion-guided weighting strategy effectively captures both geometric and semantic asymmetries in panoramic video, offering clearer training signals.

\begin{figure}[!t]
    \centering
    \includegraphics[width=\textwidth]{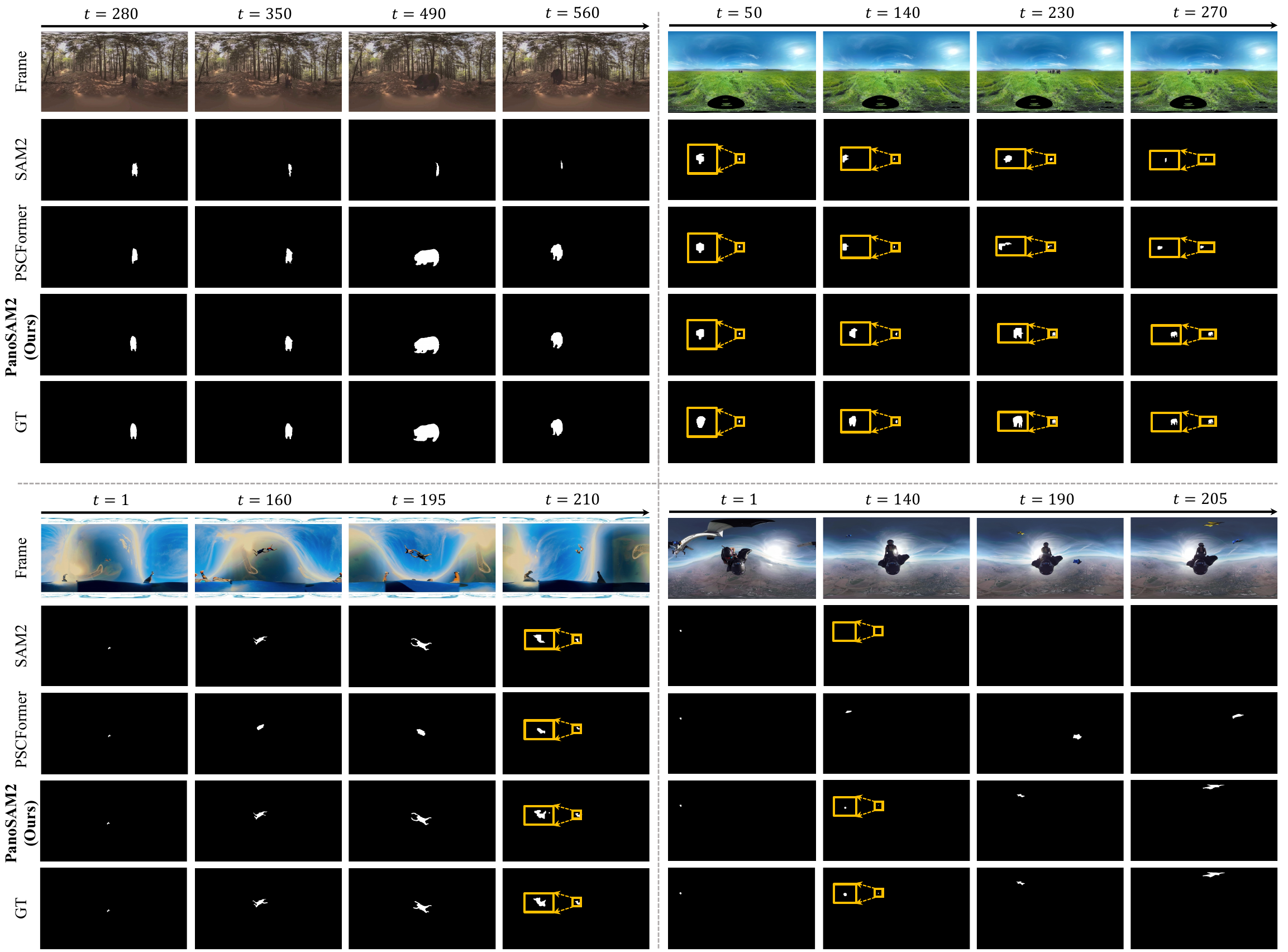}
    \caption{\textbf{More qualitative comparison between PanoSAM2 and other VOS models.} Orange bounding boxes highlight and zoom in on the small mask region.}
    \label{fig:supply_comparision}
\end{figure}

\subsection{More Qualitative Comparisons}
\cref{fig:supply_comparision} presents additional qualitative comparisons between PanoSAM2 and representative VOS models on the PanoVOS test dataset. The examples include a bear moving through a forest, a distant elephant on an open grassland, and skydivers captured from an aerial panoramic view, covering both simple and highly challenging environments. In the forest scene, PanoSAM2 consistently maintains object integrity and accurately preserves fine boundaries, whereas other methods exhibit fragmentation or drift. For the distant elephant, our model is able to track the small and low-resolution target, while competing methods struggle with missing or unstable predictions, as highlighted by the zoomed-in orange boxes. In the skydiving sequence, characterized by fast motion and extreme distortion, PanoSAM2 demonstrates strong robustness and maintains clear temporal consistency. \textbf{These visual results and Fig. 5 of the main paper collectively show that PanoSAM2 delivers superior segmentation quality across varying object scales, distortions, left–right semantic inconsistency at the 0°/360° seam, and sparse target patterns}, outperforming existing perspective and panorama-adapted VOS approaches.

\begin{figure*}[!t]
    \centering
    \includegraphics[width=\textwidth]{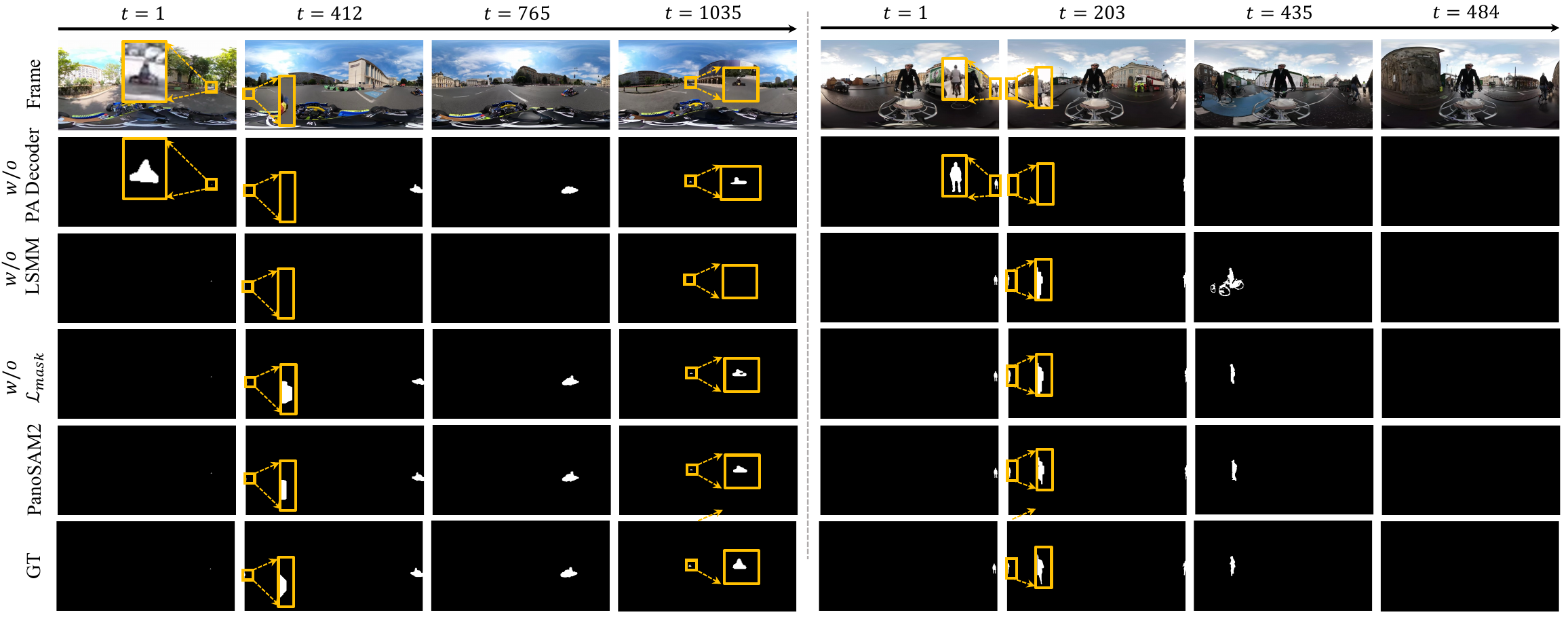}
    \caption{More ablation visualizations of proposed modules. Orange bounding boxes highlight and zoom in on the small mask region.}
    \label{fig:supply_ablation}
\end{figure*}

\subsection{More Ablation Visualizations}
We provide additional visualization of the designed component ablation in \cref{fig:supply_ablation}. The PA decoder maintains seam consistency, ensuring smooth tracking across the 0/360-degree boundary, while the LSMM module effectively prevents object drift. Additionally, the distortion-aware loss function $\mathcal{L}_{mask}$ contributes to producing a mask with more precise boundaries, further validating the effectiveness of our proposed components.

\begin{figure*}[!t]
    \centering
    \includegraphics[width=\textwidth]{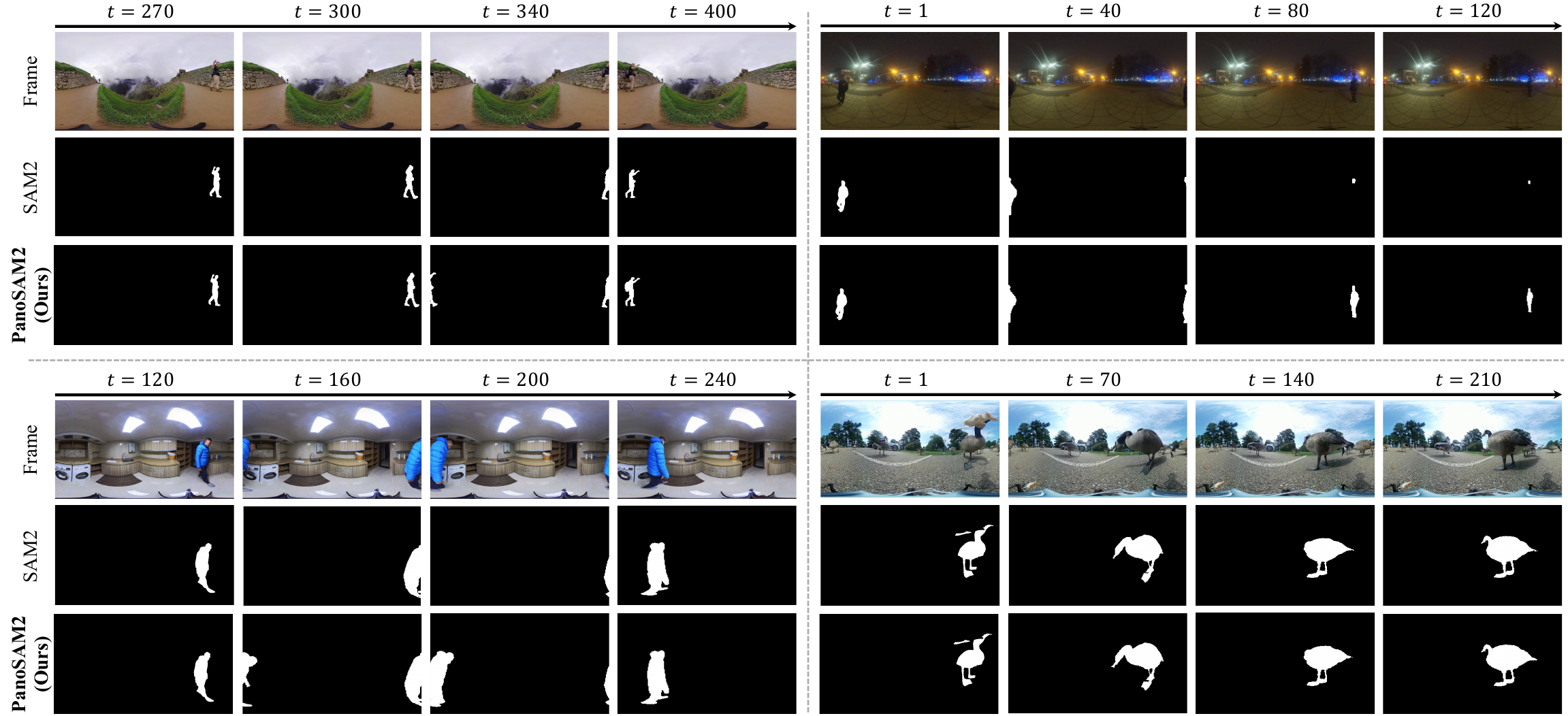}
    \caption{More visual results on self-collected open-world 360 scene.}
    \label{fig:supply_open_world}
\end{figure*}

\subsection{More Open-World Visualizations}
\cref{fig:supply_open_world} showcases additional open-world visualizations comparing PanoSAM2 with the original SAM2 on our self-collected 360 videos from many other datasets \cite{yan2024panovos, chen2024x360, tan2024imagine360}. The sequences span a wide variety of real-world environments, including both indoor and outdoor scenes, daytime and nighttime lighting, and diverse target categories such as humans and animals. Across these scenarios, PanoSAM2 consistently produces cleaner, more stable, and more complete masks than SAM2, especially in frames with strong distortion, low illumination, or fast motion. In the indoor setting, PanoSAM2 accurately preserves human contours and avoids the fragmentation observed in SAM2. In outdoor nighttime scenes, our model maintains coherent tracking despite challenging lighting, while SAM2 often loses the target. For animal sequences captured in natural environments, PanoSAM2 provides precise segmentation even when the object undergoes large pose changes. These results demonstrate that PanoSAM2 inherits the strong generalization ability of SAM2 while further improving robustness to the unique challenges of open-world 360 video.

\end{document}